\documentclass{article}

\usepackage{arxiv}

\usepackage[utf8]{inputenc} 
\usepackage[T1]{fontenc}    
\usepackage{hyperref}       
\usepackage{url}            
\usepackage{booktabs}       
\usepackage{amsfonts}       
\usepackage{nicefrac}       
\usepackage{microtype}      
\usepackage{lipsum}		
\usepackage{graphicx}
\usepackage[numbers,square]{natbib}
\usepackage{doi}

\usepackage{caption} 
\usepackage{booktabs}

\usepackage{multirow}       
\usepackage{amsmath}        
\usepackage{nicefrac}       
\usepackage{microtype}      

\usepackage{algorithm}  %
\usepackage{algpseudocode} %
\usepackage{amsthm}   
\usepackage{graphicx} %
\usepackage{mathtools}
\usepackage{bbm}
\usepackage{cleveref}
\usepackage{subcaption}
\usepackage{dsfont}
\usepackage{tcolorbox}

\usepackage{tikz,pgfplots}
\pgfplotsset{compat=1.18}

\usepackage{amssymb}

\title{Forgetting: A New Mechanism Towards Better Large Language Model Fine-tuning}

\date{} 					

\author{%
\textbf{Ali Taheri}$^{* 1}$ \quad
\textbf{Alireza Taban}$^{* 1}$ \quad
\textbf{Qizhou Wang}$^{2}$ \quad
\textbf{Shanshan Ye}$^{\dagger 3}$ \quad
\textbf{Abdolreza Mirzaei}$^{4}$ \quad
\textbf{Tongliang Liu}$^{5}$ \\
\textbf{Bo Han}$^{6, 2}$
}

\renewcommand{\shorttitle}{}
\renewcommand{\undertitle}{}
\renewcommand{\headeright}{}

\begin{document}
\maketitle
\renewcommand{\thefootnote}{}
\footnotetext{$^*$ Equal Contribution. $^\dagger$ Correspondence to Shanshan Ye. 
$^1$ Department of Electrical and Computer Engineering, Isfahan University of Technology. 
$^2$ RIKEN Center for Advanced Intelligence Project (AIP). 
$^3$ Australian Artificial Intelligence Institute, University of Technology Sydney. 
$^4$ School of Computer Science, Simon Fraser University. 
$^5$ Sydney AI Centre, The University of Sydney. 
$^6$ Department of Computer Science, Hong Kong Baptist University.}

\begin{abstract}
Supervised fine-tuning (SFT) plays a critical role for pretrained large language models (LLMs), notably enhancing their capacity to acquire domain-specific knowledge while preserving or potentially augmenting their general-purpose capabilities. However, the efficacy of SFT hinges on data quality as well as data volume, otherwise it may result in limited performance gains or even degradation relative to the associated baselines. 
To mitigate such reliance, we suggest categorizing tokens within each corpus into two parts---\textbf{positive} and \textbf{negative} tokens---based on whether they are useful to improve model performance.  
Positive tokens can be trained in common ways, whereas negative tokens, which may lack essential semantics or be misleading, should be explicitly forgotten. Overall, the token categorization facilitates the model to learn less informative messages, and the forgetting guides the model on what information to learn more precisely.
We conduct experiments across diverse and well-established benchmarks using various model architectures, demonstrating that this forgetting mechanism enhances model performance.
\end{abstract}

\section{Introduction}

In recent years, we have witnessed emerging advancements in large language models (LLMs)~\cite{brown2020languagemodels, achiam2023gpt4}, powered by transformer-based architectures~\cite{vaswani2017attention} with billions of parameters and extensive pre-training on trillions of tokens~\cite{zhao2023survey}. These models have evolved rapidly with continuous improvements in architectural design, training strategies, and scaling techniques~\cite{hoffmann2022trainingcomputeoptimal}. They exhibit exceptional performance across a wide range of complex linguistic tasks, including reasoning, solving mathematics~\cite{shao2024deepseekmath}, summarization~\cite{nallapati2016abstractive}, language understanding, code generation~\cite{chen2021evaluating,jiang2023survey}, question answering~\cite{rajpurkar2016squad}, etc.

Although powerful, LLMs still require SFT to enhance their performance in specialized tasks~\cite{chung2023scaling,aggarwal2024maple,strangmann2024transfer,lialin2023scaling}. SFT typically involves adapting the current LLM using conditional maximum likelihood principles on fine-tuning data comprising prompt-response pairs. However, its success heavily relies on the quality and volume of the data: Low quality can mislead the model learning~\cite{dodge2021documenting,luccioni2021whats,welbl2021challenges,longpre2023pretrainers}, introducing biases or inaccuracies that degrade performance, and small-scale datasets will hinder the model ability to generalize well~\cite{ghosh2024closer}. 
On the other hand, collecting the ideal data needed for SFT can be challenging in practice. Generally speaking, task-specific data are often scarce, particularly in niche or emerging domains~\cite{ghosh2024closer,ma2024investigating}, making it difficult to collect a sufficiently diverse dataset. Additionally, ensuring data quality is a non-trivial task, as it involves curating examples that are both representative and free from noise or errors. Even for humans, identifying whether the data meet high-quality standards can be difficult due to the subtleties of language and context. 
Consequently, the lack of high-quality, task-specific data becomes a bottleneck for SFT, limiting the potential of LLMs to excel in specialized applications.

\textbf{How can we mitigate the impacts of data on fine-tuning?} Data filtering~\cite{albalak2024survey} offers a promising solution. Specifically, it involves selecting a subset of data from the whole set that is expected to be more beneficial for the targeted LLM than the original.
With proper selection rules, such as gradient behaviors~\cite{albalak2023improving}, margins, loss, and influence~\cite{bejan2023make}, filtering can refine data quality effectively. However, this comes at the cost of reducing the scale of the dataset, raising open questions about the trade-off between quality and scales and its impact on the generalization of the resulting model. 
Existing literature has attempted to mitigate this issue by exploring data rephrasing~\cite{eldan2023whos,jin2024rwku}, while this approach heavily depends on manual efforts and/or expensive generators that are task-specific.

In this paper, we explore a new mechanism towards better LLM fine-tuning, referred to as \textbf{forgetting}. Following previous wisdom~\cite{yuan2024closer,eldan2023whos,wang2025rethinking,koh2017understanding}, we begin by performing data filtering at the token level, categorizing tokens as either \textbf{positive} or \textbf{negative} based on their influence to enhancing performance. Note that token-level filtering helps preserve the data scale as much as possible, thus adopting as a default choice. Then, for positive tokens, conditional maximum likelihood is applied as usual, since our selection rules ensure that their learning will benefit the current model. 
Furthermore, for negative tokens, rather than simply discarding them, we propose applying forgetting (also referred to as unlearning~\cite{li2025machine,de2021editing,jang2022knowledge,maini2024tofu,yao2024large,wang2025rethinking}) to reduce the likelihood of their generation. Compared to positive ones, negative tokens are more likely to carry uninformative or even misleading knowledge. Explicitly forgetting these tokens not only prevents the model from generating them but also helps avoid overfitting to the current corpus. Moreover, we maintain the same data scale as in conventional fine-tuning, while taking some data (tokens more accurately) as negative samples during training to facilitate model generalization.

Although straightforward to implement, we demonstrate the importance of forgetting in SFT for improved generalization through our extensive experiments. Specifically, we build our training corpus across 5 representative reasoning, knowledge and conversational datasets, and evaluate our forgetting mechanism alongside baseline methods on 5 diverse benchmark datasets, incorporating various LLMs as base models. For example, as shown in Table~\ref{tab:results} in Section~\ref{experiment}, using LLaMA3.2-1B as the base model, our approach achieved a 2.51\% improvement over that without forgetting and a 4.49\% improvement over the fine-tuned model on full tokens. Similarly, with LLaMA3.2-3B, we obtained a 3.4\% improvement over that without forgetting and 5.28\% over fine-tuned model on full tokens. Additionally, with LLaMA3.1-8B, our approach resulted in a 4.21\% improvement over the no forgetting approach, and a 8.25\% improvement over the fine-tuned model on full tokens. To validate scalability to larger model sizes, we conducted experiments with LLaMA-2-13B in Appendix~\ref{app:llama2-13b}, confirming the forgetting mechanism's generalization capability across different scales. Furthermore, we demonstrate our effectiveness across other model architectures (Qwen2.5-3B and GPT-Neo-2.7B) and diverse evaluation tasks in Appendix~\ref{app:diverse-models}.

\textbf{Connection with broader literature.} 
The mechanism of forgetting is closely connected to preference optimization (PO)~\cite{rafailov2023direct}. Recalling that, many representative PO methods, such as direct preference optimization (DPO) \cite{rafailov2023direct} and proximal policy optimization (PPO) \cite{schulman2017proximal}, can broadly be reviewed as combining the objectives of learning and forgetting. They aim to increase the likelihood of generating preferred corpora while reducing that of the dispreferred one. 
However, these methods are derived from the original PO objectives, which are inherently tied to problem setups and rely on manual labeling or reward models for preference annotation.
In contrast, we focus on the SFT problems, where the forgetting mechanism acts as an enhancement strategy rather than a indispensable component of the problem formulation.
Our method is inspired by PO but more focuses on the mechanism of forgetting as an integral component within learning. This approach helps mitigate the negative effects of low-quality data meanwhile enhancing generalization. In the long term, we aim to bridge the methodological gap between SFT and PO, striving for a more unified and flexible framework for adapting LLMs.

\section{Related works}
\label{rel_works}

\subsection{Data selection for SFT}
SFT is a well-known fine-tuning technique that maximizes the likelihood of generating target tokens under the assumption that all tokens are informative. However, data quality has emerged as a critical bottleneck for this approach \cite{luo2024robustft}, with errors arising from various sources including human annotators, tool annotators, LLM hallucinations, and data processing inconsistencies \cite{luo2024robustft}. 

LIMA \cite{zhou2023lima}, hypothesized that LLMs primarily learn the style of dataset responses, rather than updating their pre-trained knowledge toward specialized tasks, by showing that fine-tuning on a 10k carefully curated dataset, they can obtain better performance than fine-tuning on a larger dataset. 

To address quality challenges, researchers have investigated the advantages of data quality over quantity, proposing selection algorithms based on quality and diversity metrics to filter misleading samples and improve instruction-following capabilities \cite{chen2023maybe,maharana2024d2,lu2024instag,wu2023self,xia2024less}. While effective at improving performance, these approaches suffer from a fundamental limitation: they operate at the sample level, discarding entire examples and thus reducing the overall data scale available for training. This creates an inevitable trade-off between quality and quantity that remains unresolved.

Several data quality metrics have been introduced, such as gradient matching \cite{zhou2023dataset}, human feedback \cite{openassistant2023} and influence function scores \cite{xia2024less}. Moreover, \cite{dai2025improving} demonstrated that naturally higher influence scores for certain tasks can introduce bias in data selection, and proposed normalizing influence scores across different tasks before iteratively selecting samples for underrepresented skills. In \cite{luo2024robustft}, authors propose a two-stage noise-robust framework that performs noise detection using multiple expert systems and then relabels the downstream task data by finding similar examples from the clean set to provide context. 
In another approach, researchers showed that selecting training samples aligned with the model's existing knowledge can improve performance by generating multiple instruction-response pairs and choosing those with the highest probability according to the target model \cite{zhang2025best}. 

Recent studies have explored various high-quality data selection algorithms for LLM fine-tuning, yet they predominantly overlook a crucial insight: even in noisy samples, some tokens still contain valuable information. By discarding entire samples, these methods inadvertently remove useful training signals. Furthermore, these approaches fail to utilize the rejected data as a learning signal.

\subsection{LLM unlearning and PO}
\label{rel_work-unlearn}
Several approaches have been proposed to remove specific information from LLM without complete retraining them from scratch, including data replacement and relabeling strategies \cite{eldan2023whos,jin2024rwku}, and knowledge editing techniques by predicting targeted parameter updates to change specific facts while preserving other knowledge \cite{de2021editing}. Gradient ascent (GA) based methods are usually used for their simplicity, which maximize the negative log-likelihood of specific token sequences\cite{jang2022knowledge,maini2024tofu,yao2024large,tian2024forget,cha2024towards,chen2023unlearning}. However, some of them lead to degradation in LLM's outputs globally and damage the overall integrity of LLMs when removing targeted knowledge \cite{chen2023unlearning,wang2024towards,zhang2024negative,lizzo2024unlearn}---called excessive unlearning, which some regularization techniques such as minimizing the KL-Div between the output distributions of the pre-trained and fine-tuned models \cite{yao2024machine} is proposed to maintain performance on retain dataset. This introduce additional computational overhead and hyperparameter sensitivity. Researchers in \cite{wang2025rethinking} introduced WGA, which applies confidence-based weights to mitigate the excessive unlearning on a controlled forgetting manner. 

In the PO field, DPO has emerged as an alternative to PPO-based alignment methods. However, PPO has been successful for its sample efficiency compared to earlier policy gradient methods, it still suffers from explicitly modeling a reward model, and complex hyperparameter tuning \cite{schulman2017proximal}. To address these challenges and making it more robust and less computationally expensive, DPO formulates the alignment objective into a maximum likelihood formulation on a preference-paired data, trying to make preferred responses more likely and dispreferred responses less likely. There are extensive studies to address the limitations of DPO \cite{ethayarajh2024kto,azar2023general,xu2024contrastive,hong2024orpo,meng2024simpo,zeng2024token}, a new approach for preference-based unlearning was proposed by \cite{maini2024tofu}, which defines the forget set as the dispreferred responses, and the preferred response contains the refusal responses like "I do not know the answer". Inspired by this research, \cite{zhang2024negative} proposed a new variant of DPO, called negative preference optimization (NPO) that uses only negative responses, disregarding the positive ones. In the \cite{wang2025rethinking} further proposed Token-level NPO (TNPO) and Weighted TNPO (WTNPO), applying unlearning at the individual token level for more precise control over knowledge removal, yet these methods were developed specifically for targeted forgetting rather than as a complement to learning during SFT.

\section{Preliminary}
\label{preliminary}
In this section, we present the foundational background essential to our work. We start by introducing SFT for autoregressive language modeling, followed by discussing the data quality issues within SFT.

\subsection{SFT}
Autoregressive language modeling, known as sequential prediction of outputs conditioned on previous context, plays a dominant role in contemporary LLMs. 
After pre-training, SFT is typically adopted to further improve LLMs for specific tasks by optimizing on task-specific instruction-response pairs. Specifically, representing a training corpus as $D=\{(X_i,Y_i)\}^N_{i=1}$, including $N$ sequence sample pairs, each pair containing $X_i$ as an input prompt and $Y_i$ as a completion response. Each prompt $X_i$ is denoted as $X_i = \{x_{i,j}\}^{m_i}_{j=1}$ with $m_i$ indicating the sequence length of the $i$-th prompt. Similarly, each $i$-th completion response with sequence length of $n_i$ is denoted as $Y_i = \{y_{i,j}\}^{n_i}_{j=1}$. In an autoregressive manner, the model learns to estimate the probability distribution $P(y_{i,j}|X_i, y_{i,:j}; \theta)$ for each token $y_{i,j}$ in the response, conditioned on the entire prompt $X_i$ and all preceding generated tokens in the response $y_{i,:j}=\{y_{i,1}, y_{i,2}, \ldots, y_{i,j-1}\}$, where $\theta$ denotes the model parameters.

The standard cross-entropy objective is typically adopted for SFT, following the formulation of
\begin{equation}
\label{ce_loss}
\mathcal{L}(\theta) = \frac{1}{|\mathcal{I}|} \sum_{(i,j)\in \mathcal{I}} - \log P(y_{i,j}|X_i, y_{i,:j}; \theta),
\end{equation}

where the index set is defined as: 
\begin{equation}
\mathcal{I} := \{(i,j)| i \in \{1,2,\ldots,N\}, j \in \{1,2,\ldots,n_i\}\},
\end{equation}

and the per-token loss function is defined as:
\begin{equation}
\ell(y_{i,j}|x_{i,:j}; \theta) := -\log P(y_{i,j}|X_i, y_{i,:j}; \theta).
\end{equation}


\subsection{Data Quality of SFT}
\label{pre-data-quality}
LLMs acquire diverse capabilities and knowledge representations through pretraining on extensive corpora. However, for utilizing them in specialized tasks, techniques such as SFT play a remarkable role in enhancing their performance by fine-tuning the LLM on the training corpus without any selection or discarding on the dataset's components \cite{pareja2024unveiling,albalak2024survey}. 

However, collecting high-quality data, representing the required specific knowledge, is crucial to prevent inaccuracies and effectively align the LLM \cite{albalak2024survey}. High-quality data collection can be challenging in practice due to several factors. Generally, task-specific data are often scarce, particularly in emerging domains. In addition, datasets are collected from various resources, often leading to inconsistent linguistic styles and quality, and errors due to the use of annotator tools, human manual annotating \cite{luo2024robustft}. Therefore, each of them can contribute noisy and misleading tokens into the dataset thus jeopardizing the optimization process, leading to poor generalization.

To mitigate the impacts of low-quality and misleading data/tokens, existing methods proposed various data selection methods to maintain beneficial and high-quality data for fine-tuning \cite{albalak2024survey}. 
More specifically, existing methods address data filtering at the data level; however, token-level filtering seems to preserve dataset scale and fine-grained information much more. 

Although progress has been made in previous studies, they discard the low-quality data during fine-tuning, which significantly reduces the original dataset scale and potentially limits the model generalization. 
This remains an open question: how to leverage the full training dataset at its original scale while improving model performance? Specifically, is it possible to not only learn from high-quality samples but also utilize misleading data/tokens to make improvements in model generalization without overfitting to noise, while maintaining the comprehensive scope of the original dataset?

\section{Method}
\label{method}

SFT is a well-established approach for aligning extensively knowledge-augmented pretrained LLMs with specialized tasks. As discussed in Section~\ref{pre-data-quality}, practical datasets make it challenging for SFT to achieve high performance, as their collection process leads to a noisy dataset that jeopardizes the optimization process through misleading gradients. While many studies have attempted to address this issue by selecting high-quality subsets from SFT training data, these approaches sacrifice dataset scale instead of taking advantage from noisy tokens. This remained an open challenge to mitigate the effect of misleading tokens in the dataset, while preserving its scale. In this study, we propose a new approach for better LLM supervised fine-tuning, based on \textbf{forgetting} mechanism. Unlike traditional data selection approaches that treat all tokens uniformly and discard low-quality data, our method explicitly distinguishes between informative (positive) and uninformative or misleading (negative) tokens at a granular level. This token level approach preserves training data scale, while utilizing the tokens' training signals more effectively. 

Specifically, actively forgetting negative tokens, rather than merely ignoring them, can significantly improve model performance by aligning better with target data, freeing up model capacity from undesired patterns, and preventing overfitting to noisy patterns. This insight particularly valuable when working with practical datasets that inevitably include noisy tokens that should be forgotten to preserve the model's general capabilities. The overall pipeline is outlined in Algorithm~\ref{forgetting_algorithm}. In the following parts, we introduce the components of our pipeline, including the data preprocessing and training objective function.

\begin{algorithm}[ht]
\caption{Forgetting}
\label{forgetting_algorithm}
\begin{algorithmic}[1]
\Require Base model $\theta$, dataset $\mathcal{D}$, proportion $\rho$, $t_{min}$, $t_{max}$
\Ensure Fine-tuned model $\theta^*$

\State // Stage 1: Reference Model Fine-tuning
\State $\theta' \gets$ fine-tune $\theta$ on sampled subset $\mathcal{D}_{ref} \subset \mathcal{D}$

\State // Stage 2: Token Quality Assessment
\State $\mathcal{I} \gets$ All token indices $(i,j)$ in $\mathcal{D}_{train}$
\For{$(i,j) \in \mathcal{I}$}
    \State $\mathit{Inf}(y_{i,j}) \gets \ell(y_{i,j}|x_{i,:j}; \theta') - \ell(y_{i,j}|x_{i,:j}; \theta)$
    \State $\mathcal{Q}(y_{i,j}) \gets -\mathit{Inf}(y_{i,j})$ \Comment{Quality score}
\EndFor

\State // Stage 3: Token Selection
\State Sort tokens by $\mathcal{Q}(y_{i,j})$ to partition into positive and negative subsets
\State $\mathcal{P} \gets \{(i,j) \in \mathcal{I} : \mathcal{Q}(y_{i,j}|x_{i,:j}; \theta, \theta') \geq \mathcal{F}_\mathcal{S}(1-\rho)\}$ \Comment{Positive tokens}
\State $\mathcal{N} \gets \mathcal{I} \setminus \mathcal{P}$ \Comment{Negative tokens}

\State // Stage 4: Training with Forgetting
\For{$step = 0$ to $total\_steps$}
    \State $\lambda(step) \gets (t_{max} - t_{min}) \cdot \frac{step}{total\_steps}$
    \State $\mathcal{L}_\mathcal{P} \gets$ Mean weighted loss over positive tokens in $\mathcal{P}$
    \State $\mathcal{L}_\mathcal{N} \gets$ Mean weighted loss over negative tokens in $\mathcal{N}$
    \State $\mathcal{L}(\theta) \gets \mathcal{L}_\mathcal{P} - \lambda(step) \cdot \mathcal{L}_\mathcal{N}$
    \State Update $\theta$ using optimizer step on $\mathcal{L}(\theta)$
\EndFor
\State \Return $\theta$
\end{algorithmic}
\end{algorithm}

\subsection{Token quality assessment}
To quantify token quality, we leverage the concept of influence functions \cite{koh2017understanding}, between the base and reference models. Given a base model with parameters $\theta$ and a reference model with parameters $\theta'$ (introduced in Section~\ref{experiment-models}), we define the cross-model influence for token $y_{i,j}$ as follows.


\begin{equation}
\mathit{Inf}(y_{i,j}|x_{i,:j}; \theta, \theta') = \ell(y_{i,j}|x_{i,:j}; \theta') - \ell(y_{i,j}|x_{i,:j}; \theta).
\end{equation}

The intuition is that tokens that become more predictable after initial training (resulting in loss reduction) represent patterns that the model has successfully learned and are likely to be informative.

The token quality score formulation is as follows:


\begin{equation}
\label{eq:score}
\mathcal{Q}(y_{i,j}|x_{i,:j}; \theta, \theta') = -\mathit{Inf}(y_{i,j}|x_{i,:j}; \theta, \theta').
\end{equation}

A positive quality score indicates that the token became more predictable on the reference model (lower loss in $\theta'$ than in $\theta$), indicating that it represents a generalizable pattern. In contrast, a negative score suggests that the token might represent noise or misleading information.

\subsection{Token selection}

As a preprocessing step, we partition the tokens into positive and negative sets based on the quality scores. We first compute quality scores for all tokens in the training corpus, then sort them in descending order to form the set $\mathcal{S}$. Given a proportion hyperparameter $\rho \in (0,1)$, we partition the tokens as follows:


\begin{align}
\mathcal{P} &= \{(i,j) \in \mathcal{I} : \mathcal{Q}(y_{i,j}|x_{i,:j}; \theta, \theta') \geq \mathcal{F}_\mathcal{S}(1-\rho)\} \\
\mathcal{N} &= \mathcal{I} \setminus \mathcal{P}
\end{align}

where $\mathcal{F}_\mathcal{S}(1-\rho)$ denotes the $(1-\rho)$-th percentile threshold in $\mathcal{S}$. The top $\rho$ proportion of tokens are considered as \textbf{positive} tokens form the $\mathcal{P}$ set, while the remaining tokens form the \textbf{negative} set $\mathcal{N}$. In practice, we found that setting $\rho$ in the range of $0.7$ to $0.8$ achieves best results in our experiments. 
Furthermore, our experiments reveal that partitioning tokens by a zero threshold score (i.e. $\mathcal{Q} > 0$ as positive tokens) negatively affects performance. This challenges the intuition that tokens with higher confidence improvement are informative and beneficial, while the others are harmful, introducing an open challenge for proposing more robust methods to identify high-quality tokens.

\subsection{Training objective}

While standard SFT algorithms maximize the likelihood over all tokens uniformly (potentially reinforcing noisy patterns that mislead optimization) and data selection methods discard the distinguished noisy data before training, our approach maintains the benefits of full-scale training while addressing quality concerns, which 
enables improvements in the model performance, 
by minimizing the likelihood of generating the noisy tokens and freeing model capacity from misleading noisy patterns.
As mentioned in the Section~\ref{rel_work-unlearn}, unlearning techniques proven to be effective to mitigate the influence of undesirable data while preserving the model utility. In our context, rather than \textbf{forgetting} some specified knowledge(e.g., copyrighted content), we forget misleading tokens through GA, effectively utilizing both positive and negative tokens. This approach enhances the model generalization while maintaining the original data scale with no information loss. 
We propose a training objective for our selective learning and \textbf{forgetting} as follows.

\begin{equation}
\label{loss}
\mathcal{L}(\theta) = \frac{\sum_{(i,j)\in \mathcal{I}} y_{i,j} \cdot \mathbb{I}_{(i,j) \in \mathcal{P}} \cdot \ell(y_{i,j}|x_{i,:j}; \theta)}{\sum_{(i,j)\in \mathcal{I}} y_{i,j} \cdot \mathbb{I}_{(i,j) \in \mathcal{P}}} - \lambda(step) \cdot \frac{\sum_{(i,j)\in \mathcal{I}} y_{i,j} \cdot \mathbb{I}_{(i,j) \in \mathcal{N}} \cdot \ell(y_{i,j}|x_{i,:j}; \theta)}{\sum_{(i,j)\in \mathcal{I}} y_{i,j} \cdot \mathbb{I}_{(i,j) \in \mathcal{N}}},
\end{equation}

where the first term represents the average weighted loss over positive tokens, and the second term represents the average weighted loss over negative tokens. We use $\lambda(step) = (t_{\max} - t_{\min}) \cdot \frac{\text{step}}{\text{total\_steps}}$ as an adaptive coefficient that scales linearly with training progress, ensuring an effective balancing of positive and negative gradients through the optimization process. Please refer to Appendix~\ref{app:lambda} for more experiments on the $\lambda$ function selection.

In this training objective, optimization initially shares goals with generalization, but their objectives later diverge. The forgetting mechanism acts as a regularization technique that pulls optimization back for generalization when their goals conflict. By using the adaptive balancing coefficient, this enables to better capture the underlying preferred data distribution rather than overfitting to the noise or merely following the pattern of low-scale high-quality data.

However, our work differs from NPO~\cite{zhang2024negative} and TNPO~\cite{wang2025rethinking} in problem setting and mechanism design. While NPO~\cite{zhang2024negative} and TNPO~\cite{wang2025rethinking} address unlearning—removing predetermined unwanted knowledge (such as private data, copyrighted content) from trained models, our method focuses on SFT, where forgetting serves as a regularization term rather than the primary objective. We use token-level influence scores to automatically identify low-quality tokens within the training corpus to respect both the dataset quantity and quality. Then, apply forgetting to 
free model capacity from misleading noisy patterns,
simultaneously learning positive tokens and forgetting negative ones. In contrast, NPO~\cite{zhang2024negative} and TNPO~\cite{wang2025rethinking} operate on predefined forget sets where unlearning itself is the goal, not a regularization mechanism for improving generalization during task adaptation.

\section{Experiments}
\label{experiment}

\subsection{Experimental setups}
\label{experimental-setups}

\subsubsection{Datasets}
\paragraph{Training data.} We constructed our training corpus by randomly sampling from five datasets, Flan\_v2 \cite{chung2023scaling}, Dolly \cite{databricks2023dolly}, Open Assistant 1 \cite{openassistant2023}, Stanford Alpaca \cite{taori2023alpaca} and WizardLM \cite{xu2023wizardlm}. 
Please refer to Appendix~\ref{app:dataset} for more datasets details.
The dataset distribution presented in detail in Table~\ref{tab:dataset-distribution}.
This corpus provides a comprehensive coverage of domains and response styles, thereby enhancing the model's generalization capabilities \cite{wang2023camels}.
\paragraph{Evaluation benchmarks.} For the evaluation part, we have performed comprehensive evaluations on five diverse benchmark datasets. They are TruthfulQA \cite{lin2022truthfulqa} to evaluate the ability of LLM in providing truthful and accurate information, BoolQ \cite{clark2019boolq} a binray question-answering dataset and evaluates LLM's ability in making precise boolean judgements, LogiQA \cite{liu2020logiqa} focused on logical reasoning, TydiQA \cite{clark2020tydiqa} to evaluate the LLM on multilingual question-answering and ASDiv \cite{miao2021diverse} to evaluate the LLM on math word problems. The benchmarks' attributes are presented in Table~\ref{eval-table}. The evaluation is processed on all benchmark samples, by using the lm-eval-hareness\footnote{\url{https://github.com/EleutherAI/lm-evaluation-harness}} repository.

\begin{table}[t]
  \caption{Dataset distribution comparison}
  \label{tab:dataset-distribution}
  \centering
  \begin{tabular}{lrrrr}
    \toprule
    \multirow{2}{*}{Dataset} & \multicolumn{2}{c}{50k Sample} & \multicolumn{2}{c}{10k Sample} \\
    \cmidrule(lr){2-3} \cmidrule(lr){4-5}
    & Samples & Percentage & Samples & Percentage \\
    \midrule
    Dolly & 2,617 & 5.23\% & 503 & 5.03\% \\
    Flan\_v2 & 17,803 & 35.61\% & 3,593 & 35.93\% \\
    Open Assistant 1 & 5,960 & 11.92\% & 1,135 & 11.35\% \\
    Stanford Alpaca & 9,276 & 18.55\% & 1,834 & 18.34\% \\
    WizardLM & 14,344 & 28.69\% & 2,935 & 29.35\% \\
    \bottomrule
  \end{tabular}
\end{table}

\subsubsection{Models}
\label{experiment-models}
\paragraph{Base models.}
In this paper, we choose 3 open-source LLMs including LLaMA-3.2-1B, LLaMA-3.2-3B and LLaMA-3.1-8B \cite{dubey2024llama3} in diverse complexity as our base models for fine-tuning.
\paragraph{Reference models.}
The reference models are obtained by fine-tuning the base models on a subset $\mathcal{D}_{\text{ref}} \subset \mathcal{D} \text{ with } \mathcal{D}_{\text{ref}} \cap \mathcal{D}_{\text{train}} = \emptyset$ where $\mathcal{D_{\text{train}}}$ is the training corpus and $\mathcal{D}$ is a combination of training datasets. The fine-tuned LLM will be used for calculating the influence scores. We also investigate the robustness of our approach when the reference dataset contains duplicate samples (see Appendix~\ref{app:reference-duplicates}).

\paragraph{Baselines.}
In this study, our baselines include the base model, the supervised fine-tuned version of the base model on the whole training dataset with full tokens, and the fine-tuned version of the base model on the preprocessed training dataset including only the top k\% clean tokens.

\subsubsection{Training configurations}
\label{training-config}
For the reported results in Table~\ref{tab:results}, we employed model-specific hyperparameter pairs $(t_{\min}, t_{\max})$ as follows: $(10^{-5}, 0.25)$ for LLaMA-3.2-1B and $(10^{-4}, 0.25)$ for both LLaMA-3.2-3B and LLaMA-3.1-8B, for our adaptive balancing coefficient $\lambda(step)$. These values were determined through ablation studies optimizing for performance across our benchmark tasks.
For fine-tuning the LLMs, we used LoRA \cite{hu2022lora} for its memory efficiency and stability during training. We set rank-size of 64, the scaling factor of 16 and dropout 0.1 for LoRA. We used the AdamW optimizer \cite{loshchilov2017decoupled}, with the overall batch size equal to 24 and the fine-tuning process is performed for 1 epoch with a learning rate $10^{-4}$ and a linear learning rate scheduler with 0.03 warm-up ratio. Moreover, we conducted our experiments on 4 NVIDIA L40S-48GB GPUs with Intel Xeon 6338 CPUs, running on Ubuntu 20.04.6 LTS. The systems utilize Transformers version 4.51.3 and CUDA version 12.5. Training time for 1B, 3B and 8B models approximately takes 2, 3, and 5 hours, respectively.

\begin{table}[t]
  \caption{Evaluation datasets attributes}
  \label{eval-table}
  \centering
  \begin{tabular}{llll}
    \toprule
    Dataset & Focus Area & Data Size & Question Length \\
    \midrule
    TruthfulQA & Truthfulness & 817 & Medium \\
    BoolQ & Boolean QA & 15,942 & Short \\
    LogiQA & Logical reasoning & 8,678 & Medium \\
    TydiQA & Multilingual QA & 204k & Varied \\
    ASDiv & Math Word Problem Solving & 2,305 & Varied \\
    \bottomrule
  \end{tabular}
\end{table}

\begin{table}[ht]
\caption{Performance comparison of different methods across five different benchmarks using LLaMA-3.2-1B, LLaMA-3.2-3B and LLaMA-3.1-8B variants as our base models. We evaluate four approaches: Base (unmodified), Full Tokens (standard SFT), Ignoring, and our proposed \textbf{Forgetting}. The results show accuracy (\%) for TruthfulQA, BoolQ, LogiQA, and ASDiv, and one-shot F1 score for TydiQA. Bold values demonstrate best performance on each benchmark. Results show mean values with standard deviations from 3 independent training runs. Our proposed Forgetting method achieves significant improvements across different benchmarks and model scales.}

\label{tab:results}
\centering

\renewcommand{\arraystretch}{1.3}
\resizebox{\textwidth}{!}{%
\begin{tabular}{lccccccc}
\hline\hline
Method & TruthfulQA & BoolQ & LogiQA & TydiQA & ASDiV & AVG \\
\hline\hline
\multicolumn{7}{c}{Base model: LLaMA-3.2-1B} \\
\hline\hline
Base & 37.83±0 & 63.80±0 & 22.17±0 & 14.36±0 & 0±0 & 27.63±0 \\
Full Tokens & 38.74±0.39 & 59.84±0.94 & 24.60±0.25 & 28.10±0.46 & 0.55±0.48 & 30.37±0.39 \\
\hline
Ignoring (seq-level) & 39.56±0.57 & 61.47±0.06 & 24.03±0.25 & 27.90±0.34 & 1.46±0.15 & 30.88±0.28 \\
Forgetting (seq-level) & 38.93±0.08 & 63.13±0.46 & 24.80±0.12 & 28.75±0.23 & 2.50±0.04 & 31.62±0.10 \\
\hline
Ignoring (token-level) & 42.40±0.13 & 60.21±1.66 & 24.34±0.31 & 33.87±0.64 & 0.91±0.2 & 32.35±0.46 \\
Forgetting (token-level) & \textbf{44.83±0.45} & \textbf{65.39±0.39} & \textbf{25.60±0.48} & \textbf{36.21±0.77} & \textbf{2.28±0.04} & \textbf{34.86±0.22} \\
\hline\hline
\multicolumn{7}{c}{Base model: LLaMA-3.2-3B} \\
\hline\hline
Base & 39.45±0 & 73.04±0 & 22.17±0 & 21.12±0 & 31.24±0 & 37.40±0 \\
Full Tokens & 42.95±0.47 & 72.54±0.59 & 25.51±0.21 & 44.04±0.27 & 49.46±0.14 & 46.90±0.16 \\
\hline
Ignoring (seq-level) & 40.58±0.54 & 72.93±0.28 & 24.36±0.47 & 44.82±1.03 & 49.11±0.29 & 46.36±0.41 \\
Forgetting (seq-level) & 40.95±0.30 & \textbf{77.80±0.38} & 25.27±0.23 & 47.52±0.45 & 49.83±0.93 & 48.27±0.06 \\
\hline
Ignoring (token-level) & 47.23±0.86 & 75.40±0.37 & 25.12±0.31 & 47.63±0.42 & 48.51±0.74 & 48.78±0.19 \\
Forgetting (token-level) & \textbf{50.32±0.96} & 76.66±0.07 & \textbf{27.09±0.37} & \textbf{56.36±0.06} & \textbf{50.47±0.3} & \textbf{52.18±0.12} \\
\hline\hline
\multicolumn{7}{c}{Base model: LLaMA-3.1-8B} \\
\hline\hline
Base & 45.08±0 & 82.15±0 & 26.51±0 & 46.67±0 & 12.93±0 & 42.67±0 \\
Full Tokens & 44.51±0.48 & 81.44±0.47 & 25.68±0.14 & 52.03±0.18 & 51.46±0.42 & 51.02±0.11 \\
\hline
Ignoring (seq-level) & 47.05±0.21 & 85.17±0.45 & 24.64±0.18 & 52.34±0.08 & 51.62±0.28 & 52.16±0.24 \\
Forgetting (seq-level) & 47.83±0.09 & \textbf{85.56±0.18} & 24.85±0.37 & 57.56±0.33 & 57.76±0.18 & 54.71±0.10 \\
\hline
Ignoring (token-level) & 52.38±0.22 & 82.76±0.07 & 25.53±0.11 & 56.66±0.06 & \textbf{57.95±0.35} & 55.06±0.16 \\
Forgetting (token-level) & \textbf{58.39±0.65} & 83.14±0.15 & \textbf{31.15±0.86} & \textbf{66.21±0.23} & 57.48±0.12 & \textbf{59.27±0.35} \\
\hline\hline
\end{tabular}
}
\end{table}

\subsection{Empirical Results}
We conducted comprehensive experiments to evaluate our forgetting approach against all baselines. Remarkably, our method outperformed all baselines in average performance. The forgetting method achieved superior results with $\rho$ in the range of 70\% to 80\%, while the ignoring has its best-case performance with $\rho$ in the range of 50\% to 60\% across all benchmarks. We demonstrate the results of our experiments utilizing three different variants of LLaMA in Table~\ref{tab:results}, comparing the method in their best-case performance, specifically, setting $\rho=0.7$ for our forgetting approach and $\rho=0.5$ for the ignoring approach. Notably, compared to the standard SFT our method has achieved an average performance improvement of $4.49\%$ on the 1B model, $5.28\%$ on the 3B model and $8.25\%$ on the 8B model. Furthermore, compared to ignoring baseline, our method has achieved performance improvement of $2.51\%$ on the 1B model, $3.4\%$ on the 3B model and $4.21\%$ on the 8B model. Please see Appendix~\ref{app:time_complexity} for a detailed analysis of computational overhead.

Additional experiments with LLaMA-2-13B \cite{touvron2023llama} confirms these forgetting mechanism's generalization capability in larger scales, with detailed results provided in Appendix~\ref{app:llama2-13b}. To further validate the generalizability of our forgetting mechanism across different model architectures and benchmarks, we conducted additional experiments on Qwen2.5-3B~\cite{yang2024qwen25} and GPT-Neo-2.7B~\cite{black2021gptneo} across four diverse benchmarks, Instruction-Following~\cite{zhou2023ifeval}, ARC-Challenge~\cite{clark2018arc}, LAMBADA~\cite{paperno2016lambada} specially using OpenAI preprocessing~\cite{radford2019language} from the EleutherAI\footnote{\url{https://huggingface.co/datasets/EleutherAI/lambada_openai}} repository, and Arithmetic~\cite{brown2020languagemodels}. The results, presented in Appendix~\ref{app:diverse-models}, demonstrate the superiority of our forgetting mechanism. Notably, our forgetting method achieved a 5.33\% improvement over the ignoring baseline on Qwen2.5-3B and a 3.56\% improvement on GPT-Neo-2.7B, confirming that the benefits of our approach extend beyond the LLaMA family and linguistics task.

\textbf{Token-level vs. sequence-level granularity.} A key design choice in our approach is operating at the token level rather than the sequence level. This granular approach is motivated by the observation that individual sequences often contain a mixture of both informative and misleading tokens. Sequence-level selection would classify entire sequences as either positive or negative, potentially discarding valuable tokens within otherwise noisy sequences, or conversely, retaining harmful tokens within generally useful sequences. Token-level selection allows us to preserve beneficial information while selectively forgetting problematic content, maximizing the utility of our training data. The Table~\ref{tab:results} shows a comparison of the different approaches.

Table~\ref{tab:results} shows that token-level approaches consistently outperform their sequence-level counterparts across all model sizes. For example, with LLaMA-3.2-3B, token-level forgetting achieves 52.18\% average performance compared to 48.27\% for sequence-level forgetting. This superiority stems from token-level selection's ability to preserve useful information even in partially noisy sequences, while sequence-level selection discards entire sequences that may contain valuable tokens alongside problematic ones.

\subsection{Ablation study}
\begin{figure}[t]
   \centering
   \begin{minipage}[b]{0.48\linewidth}
       \centering
       \includegraphics[width=\linewidth]{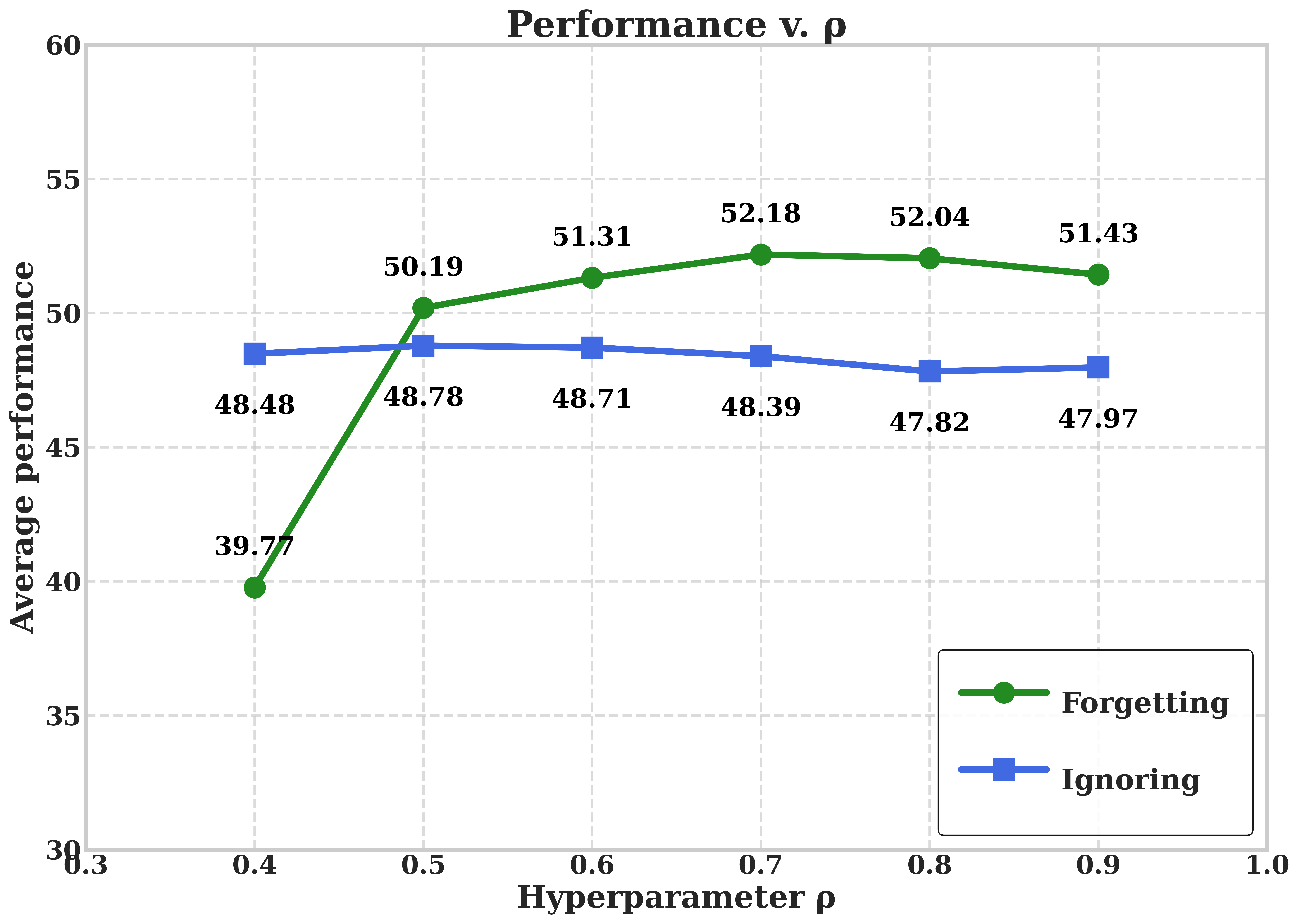}
       \centerline{(a)}
   \end{minipage}
   \hfill
   \begin{minipage}[b]{0.48\linewidth}
       \centering
       \includegraphics[width=\linewidth]{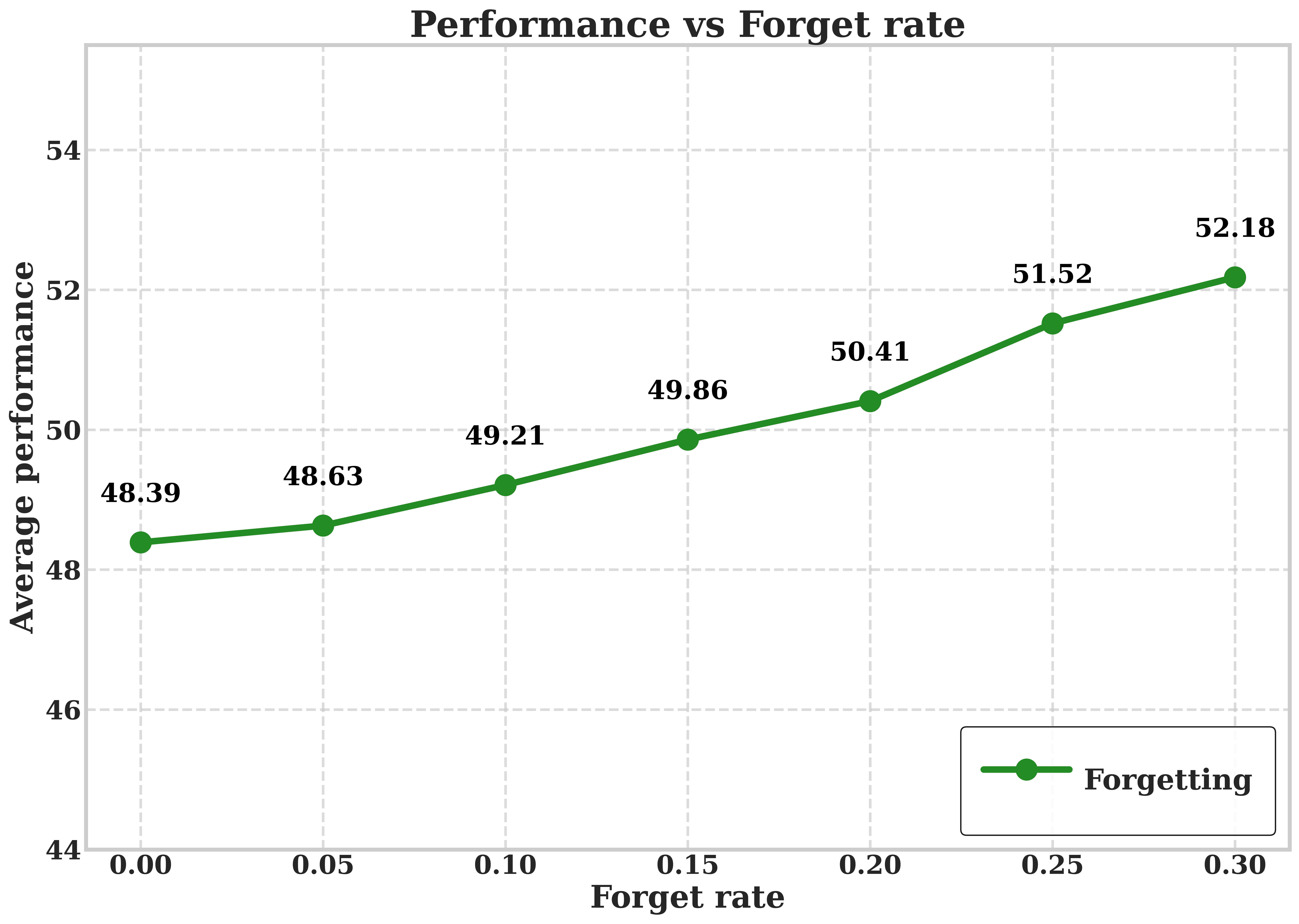}
       \centerline{(b)}
   \end{minipage}
   
   \caption{Performance analysis: (a) Average performance of forgetting versus ignoring methods across different $\rho$ values. (b) Average performance of the forgetting method with different forget rates.}
   \label{fig:performance_analysis}
\end{figure}

\textbf{Impact of $\rho$.}  Our empirical evidence indicates that the forgetting approach demonstrates superior generalization capability when $\rho$ has a higher value, partitioning a larger subset of tokens as positive tokens and treating all remaining tokens as negative tokens (forget rate of $1-\rho$). However, forgetting only a subset of the remaining tokens and discarding the others leads to suboptimal performance, indicating the effectiveness of forgetting all the $1-\rho$ tokens as negative tokens. Figure~\ref{fig:performance_analysis}(b) illustrates the average performance for different forget rates. Moreover, the choice of the hyperparameter $\rho$, directly affects the noise distribution in positive and negative sets. Higher value of $\rho$ can introduce noisy tokens to the positive set, while lower value of $\rho$ can add informative tokens to the negative set.  Figure~\ref{fig:performance_analysis}(a) shows the comparison between different values of $\rho$ for the forgetting and ignoring approaches. The average performance of the forgetting method has significantly decreased for the lowest value $\rho=0.4$, due to the higher proportion of informative tokens in the negative set.\\
\textbf{Impact of $\lambda$(step).} As explained in Section~\ref{method}, effectively balancing the training and forgetting gradients is crucial for optimization stability. As related studies typically use a constant coefficient in the range (0,1) to reduce the learning rate of forgetting gradients. However, through empirical investigation, we observed that as training iterations progress, the learning rate reduction leads to the vanishing of the forgetting gradients. Thus, we used an adaptive function $\lambda(step)$, as a coefficient on forgetting loss term of our dual objective function, not only to balance the learning and forgetting gradients, but also to efficiently preserve the effects of forgetting gradients during fine-tuning. According to the dual objective function formula, ignoring approach is equivalent to forgetting with a balancing coefficient of zero. 
In a comparison of balancing coefficient strategies, we evaluated three approaches: static approaches with constant values zero (ignoring) and 0.0001 (optimal value for static strategy), and a dynamic approach using the linear function $\lambda(step)$ with $t_{min}=0.0001$ and $t_{max}=0.25$. The corresponding average improvements are 48.78\%, 49.59\%, and 52.18\%, respectively. These results demonstrate that adaptive adjustment via linear function significantly outperforms static coefficient assignment, highlighting the critical role of selecting an appropriate balancing coefficient strategy. By incorporating $\lambda(step)$, the forgetting learning rate decreases more gradually with a shallower slope. We investigated the impact of the adaptive parameter $\lambda(step)$ through a series of experiments. \\
\textbf{Hyperparameter sensitivity analysis.} To evaluate the robustness of our approach to hyperparameter choices, we conducted extensive experiments varying the key parameters $t_{\min}$ and $t_{\max}$ while keeping $\rho=0.7$ fixed. As shown in Figure~\ref{fig:performance_analysis}(a), our method demonstrates impressive robustness to $\rho$ values across a wide range. For practical selection of $\rho$, users can use the ratio of tokens with positive influence scores as an initial estimate—in our experiments, this ratio was 0.67, leading us to select $\rho=0.7$ as optimal. Comprehensive results across different combinations of $t_{\min}$ and $t_{\max}$ values using LLaMA-3.2-3B are presented in Appendix~\ref{app:hyperparameter-sensitivity}.\\
\textbf{Impact of forgetting.} As demonstrated in previous sections, the forgetting mechanism significantly improves the performance of fine-tuning with respect to that without forgetting and standard SFT. Specifically, when comparing the forgetting and ignoring approaches with the same selection ratio ($\rho$ = 0.7), the forgetting method achieves an accuracy of 52.18\%, outperforming the ignoring approach (48.39\%). This performance gap indicates that the negative tokens set has a high noise ratio, reinforcing the impact of forgetting misleading tokens, leading to higher performance.

\section{Limitations}
\label{limitations}
Despite our method's improvements, some limitations remain. The approach is sensitive to dataset size and noise ratio, leading to performance degradation for smaller negative token sets. However, it is worth noting that noise existence is common in real-world practical datasets. Additionally, computational budget restricted our experiments to models up to 13B parameters with limited-scale training data. The performance remains uncertain how well the mechanism would perform on larger-scale base models and datasets.

\section{Conclusion}
\label{conclusion}
This paper aims to reduce the reliance of LLM fine-tuning on  data quality, an important and on-going topic that has been receiving increasing attentions these days. 
Unlike previous works that primarily focus on improving data selection, we suggest that exploring new learning paradigms is equally crucial. Specifically, we propose a novel fine-tuning mechanism named {forgetting}, which explicitly enables the model to forget misleading message carried by those filtered-out tokens. It mitigates the negative impact of noisy or misleading data while preserving the dataset scale, 
helping to improve 
generalization and overall performance. In the future, we will explore more formal and rigorous ways to defining and enhancing data quality, as well as extend the forgetting mechanism to other related areas within LLMs, such as pre-training, preference optimization, and inference.

\bibliographystyle{unsrtnat}  
\bibliography{references}      

@inproceedings{brown2020languagemodels,
	author = {Brown, Tom and Mann, Benjamin and Ryder, Nick and Subbiah, Melanie and Kaplan, Jared D and Dhariwal, Prafulla and Neelakantan, Arvind and Shyam, Pranav and Sastry, Girish and Askell, Amanda and Agarwal, Sandhini and Herbert-Voss, Ariel and Krueger, Gretchen and Henighan, Tom and Child, Rewon and Ramesh, Aditya and Ziegler, Daniel and Wu, Jeffrey and Winter, Clemens and Hesse, Chris and Chen, Mark and Sigler, Eric and Litwin, Mateusz and Gray, Scott and Chess, Benjamin and Clark, Jack and Berner, Christopher and McCandlish, Sam and Radford, Alec and Sutskever, Ilya and Amodei, Dario},
	booktitle = {Advances in Neural Information Processing Systems},
	editor = {H. Larochelle and M. Ranzato and R. Hadsell and M.F. Balcan and H. Lin},
	pages = {1877--1901},
	publisher = {Curran Associates, Inc.},
	title = {Language Models are Few-Shot Learners},
	url = {https://proceedings.neurips.cc/paper_files/paper/2020/file/1457c0d6bfcb4967418bfb8ac142f64a-Paper.pdf},
	volume = {33},
	year = {2020}
}

@article{achiam2023gpt4,
	title={Gpt-4 technical report},
	author={Achiam, Josh and Adler, Steven and Agarwal, Sandhini and Ahmad, Lama and Akkaya, Ilge and Aleman, Florencia Leoni and Almeida, Diogo and Altenschmidt, Janko and Altman, Sam and Anadkat, Shyamal and others},
	journal={arXiv preprint arXiv:2303.08774},
	year={2023}
}

@inproceedings{vaswani2017attention,
	author = {Vaswani, Ashish and Shazeer, Noam and Parmar, Niki and Uszkoreit, Jakob and Jones, Llion and Gomez, Aidan N and Kaiser, \L ukasz and Polosukhin, Illia},
	booktitle = {Advances in Neural Information Processing Systems},
	editor = {I. Guyon and U. Von Luxburg and S. Bengio and H. Wallach and R. Fergus and S. Vishwanathan and R. Garnett},
	pages = {},
	publisher = {Curran Associates, Inc.},
	title = {Attention is All you Need},
	url = {https://proceedings.neurips.cc/paper_files/paper/2017/file/3f5ee243547dee91fbd053c1c4a845aa-Paper.pdf},
	volume = {30},
	year = {2017}
}

@article{zhao2023survey,
	title={A survey of large language models},
	author={Zhao, Wayne Xin and Zhou, Kun and Li, Junyi and Tang, Tianyi and Wang, Xiaolei and Hou, Yupeng and Min, Yingqian and Zhang, Beichen and Zhang, Junjie and Dong, Zican and others},
	journal={arXiv preprint arXiv:2303.18223},
	volume={1},
	number={2},
	pages={1--124},
	year={2023}
}

@article{hoffmann2022trainingcomputeoptimal,
	title={Training compute-optimal large language models},
	author={Hoffmann, Jordan and Borgeaud, Sebastian and Mensch, Arthur and Buchatskaya, Elena and Cai, Trevor and Rutherford, Eliza and Casas, DDL and Hendricks, Lisa Anne and Welbl, Johannes and Clark, Aidan and others},
	journal={arXiv preprint arXiv:2203.15556},
	volume={10},
	year={2022}
}

@inproceedings{dodge2021documenting,
	title = "Documenting Large Webtext Corpora: A Case Study on the Colossal Clean Crawled Corpus",
	author = "Dodge, Jesse  and
	Sap, Maarten  and
	Marasovi{\'c}, Ana  and
	Agnew, William  and
	Ilharco, Gabriel  and
	Groeneveld, Dirk  and
	Mitchell, Margaret  and
	Gardner, Matt",
	editor = "Moens, Marie-Francine  and
	Huang, Xuanjing  and
	Specia, Lucia  and
	Yih, Scott Wen-tau",
	booktitle = "Proceedings of the 2021 Conference on Empirical Methods in Natural Language Processing",
	month = nov,
	year = "2021",
	address = "Online and Punta Cana, Dominican Republic",
	publisher = "Association for Computational Linguistics",
	url = "https://aclanthology.org/2021.emnlp-main.98/",
	doi = "10.18653/v1/2021.emnlp-main.98",
	pages = "1286--1305"
}

@inproceedings{welbl2021challenges,
	title = "Challenges in Detoxifying Language Models",
	author = "Welbl, Johannes  and
	Glaese, Amelia  and
	Uesato, Jonathan  and
	Dathathri, Sumanth  and
	Mellor, John  and
	Hendricks, Lisa Anne  and
	Anderson, Kirsty  and
	Kohli, Pushmeet  and
	Coppin, Ben  and
	Huang, Po-Sen",
	editor = "Moens, Marie-Francine  and
	Huang, Xuanjing  and
	Specia, Lucia  and
	Yih, Scott Wen-tau",
	booktitle = "Findings of the Association for Computational Linguistics: EMNLP 2021",
	month = nov,
	year = "2021",
	address = "Punta Cana, Dominican Republic",
	publisher = "Association for Computational Linguistics",
	url = "https://aclanthology.org/2021.findings-emnlp.210/",
	doi = "10.18653/v1/2021.findings-emnlp.210",
	pages = "2447--2469"
}

@inproceedings{luccioni2021whats,
	title = "What{'}s in the Box? An Analysis of Undesirable Content in the {C}ommon {C}rawl Corpus",
	author = "Luccioni, Alexandra  and
	Viviano, Joseph",
	editor = "Zong, Chengqing  and
	Xia, Fei  and
	Li, Wenjie  and
	Navigli, Roberto",
	booktitle = "Proceedings of the 59th Annual Meeting of the Association for Computational Linguistics and the 11th International Joint Conference on Natural Language Processing (Volume 2: Short Papers)",
	month = aug,
	year = "2021",
	address = "Online",
	publisher = "Association for Computational Linguistics",
	url = "https://aclanthology.org/2021.acl-short.24/",
	doi = "10.18653/v1/2021.acl-short.24",
	pages = "182--189"
}

@inproceedings{longpre2023pretrainers,
	title = "A Pretrainer{'}s Guide to Training Data: Measuring the Effects of Data Age, Domain Coverage, Quality, {\&} Toxicity",
	author = "Longpre, Shayne  and
	Yauney, Gregory  and
	Reif, Emily  and
	Lee, Katherine  and
	Roberts, Adam  and
	Zoph, Barret  and
	Zhou, Denny  and
	Wei, Jason  and
	Robinson, Kevin  and
	Mimno, David  and
	Ippolito, Daphne",
	editor = "Duh, Kevin  and
	Gomez, Helena  and
	Bethard, Steven",
	booktitle = "Proceedings of the 2024 Conference of the North American Chapter of the Association for Computational Linguistics: Human Language Technologies (Volume 1: Long Papers)",
	month = jun,
	year = "2024",
	address = "Mexico City, Mexico",
	publisher = "Association for Computational Linguistics",
	url = "https://aclanthology.org/2024.naacl-long.179/",
	doi = "10.18653/v1/2024.naacl-long.179",
	pages = "3245--3276"
}

@article{shao2024deepseekmath,
	title={Deepseekmath: Pushing the limits of mathematical reasoning in open language models},
	author={Shao, Zhihong and Wang, Peiyi and Zhu, Qihao and Xu, Runxin and Song, Junxiao and Bi, Xiao and Zhang, Haowei and Zhang, Mingchuan and Li, YK and Wu, Yang and others},
	journal={arXiv preprint arXiv:2402.03300},
	year={2024}
}

@inproceedings{nallapati2016abstractive,
	title = "Abstractive Text Summarization using Sequence-to-sequence {RNN}s and Beyond",
	author = "Nallapati, Ramesh  and
	Zhou, Bowen  and
	dos Santos, Cicero  and
	Gu{\ensuremath{\dot{}}}l{\c{c}}ehre, {\c{C}}a{\u{g}}lar  and
	Xiang, Bing",
	editor = "Riezler, Stefan  and
	Goldberg, Yoav",
	booktitle = "Proceedings of the 20th {SIGNLL} Conference on Computational Natural Language Learning",
	month = aug,
	year = "2016",
	address = "Berlin, Germany",
	publisher = "Association for Computational Linguistics",
	url = "https://aclanthology.org/K16-1028/",
	doi = "10.18653/v1/K16-1028",
	pages = "280--290"
}

@article{yuan2024closer,
	title={A closer look at machine unlearning for large language models},
	author={Yuan, Xiaojian and Pang, Tianyu and Du, Chao and Chen, Kejiang and Zhang, Weiming and Lin, Min},
	journal={arXiv preprint arXiv:2410.08109},
	year={2024}
}

@inproceedings{ghosh2024closer,
	author = {Ghosh, Sreyan and Evuru, Chandra Kiran Reddy and Kumar, Sonal and S, Ramaneswaran and Aneja, Deepali and Jin, Zeyu and Duraiswami, Ramani and Manocha, Dinesh},
	title = {A closer look at the limitations of instruction tuning},
	year = {2024},
	publisher = {JMLR.org},
	booktitle = {Proceedings of the 41st International Conference on Machine Learning},
	articleno = {624},
	numpages = {31},
	location = {Vienna, Austria},
	series = {ICML'24}
}

@article{ma2024investigating,
	title={Investigating public fine-tuning datasets: A complex review of current practices from a construction perspective},
	author={Ma, Runyuan and Li, Wei and Shang, Fukai},
	journal={arXiv preprint arXiv:2407.08475},
	year={2024}
}

@inproceedings{rajpurkar2016squad,
	title = "{SQ}u{AD}: 100,000+ Questions for Machine Comprehension of Text",
	author = "Rajpurkar, Pranav  and
	Zhang, Jian  and
	Lopyrev, Konstantin  and
	Liang, Percy",
	editor = "Su, Jian  and
	Duh, Kevin  and
	Carreras, Xavier",
	booktitle = "Proceedings of the 2016 Conference on Empirical Methods in Natural Language Processing",
	month = nov,
	year = "2016",
	address = "Austin, Texas",
	publisher = "Association for Computational Linguistics",
	url = "https://aclanthology.org/D16-1264/",
	doi = "10.18653/v1/D16-1264",
	pages = "2383--2392"
}

@article{chung2023scaling,
	author  = {Hyung Won Chung and Le Hou and Shayne Longpre and Barret Zoph and Yi Tay and William Fedus and Yunxuan Li and Xuezhi Wang and Mostafa Dehghani and Siddhartha Brahma and Albert Webson and Shixiang Shane Gu and Zhuyun Dai and Mirac Suzgun and Xinyun Chen and Aakanksha Chowdhery and Alex Castro-Ros and Marie Pellat and Kevin Robinson and Dasha Valter and Sharan Narang and Gaurav Mishra and Adams Yu and Vincent Zhao and Yanping Huang and Andrew Dai and Hongkun Yu and Slav Petrov and Ed H. Chi and Jeff Dean and Jacob Devlin and Adam Roberts and Denny Zhou and Quoc V. Le and Jason Wei},
	title   = {Scaling Instruction-Finetuned Language Models},
	journal = {Journal of Machine Learning Research},
	year    = {2024},
	volume  = {25},
	number  = {70},
	pages   = {1--53},
	url     = {http://jmlr.org/papers/v25/23-0870.html}
}

@inproceedings{aggarwal2024maple,
	title = "{MAPLE}: Multilingual Evaluation of Parameter Efficient Finetuning of Large Language Models",
	author = "Aggarwal, Divyanshu  and
	Sathe, Ashutosh  and
	Watts, Ishaan  and
	Sitaram, Sunayana",
	editor = "Ku, Lun-Wei  and
	Martins, Andre  and
	Srikumar, Vivek",
	booktitle = "Findings of the Association for Computational Linguistics: ACL 2024",
	month = aug,
	year = "2024",
	address = "Bangkok, Thailand",
	publisher = "Association for Computational Linguistics",
	url = "https://aclanthology.org/2024.findings-acl.881/",
	doi = "10.18653/v1/2024.findings-acl.881",
	pages = "14824--14867"
}

@article{strangmann2024transfer,
	title={Transfer learning for finetuning large language models},
	author={Strangmann, Tobias and Purucker, Lennart and Franke, J{\"o}rg KH and Rapant, Ivo and Ferreira, Fabio and Hutter, Frank},
	journal={arXiv preprint arXiv:2411.01195},
	year={2024}
}

@article{albalak2024survey,
	title={A survey on data selection for language models},
	author={Albalak, Alon and Elazar, Yanai and Xie, Sang Michael and Longpre, Shayne and Lambert, Nathan and Wang, Xinyi and Muennighoff, Niklas and Hou, Bairu and Pan, Liangming and Jeong, Haewon and others},
	journal={arXiv preprint arXiv:2402.16827},
	year={2024}
}

@inproceedings{albalak2023improving,
	author = {Albalak, Alon and Raffel, Colin A and Wang, William Yang},
	booktitle = {Advances in Neural Information Processing Systems},
	editor = {A. Oh and T. Naumann and A. Globerson and K. Saenko and M. Hardt and S. Levine},
	pages = {24754--24780},
	publisher = {Curran Associates, Inc.},
	title = {Improving Few-Shot Generalization by Exploring and Exploiting Auxiliary Data},
	url = {https://proceedings.neurips.cc/paper_files/paper/2023/file/4e3c5399729e06d2f0c22d04416904ab-Paper-Conference.pdf},
	volume = {36},
	year = {2023}
}

@article{maharana2024d2,
	title={D2 pruning: Message passing for balancing diversity and difficulty in data pruning},
	author={Maharana, Adyasha and Yadav, Prateek and Bansal, Mohit},
	journal={arXiv preprint arXiv:2310.07931},
	year={2023}
}

@inproceedings{bejan2023make,
	title = "Make Every Example Count: On the Stability and Utility of Self-Influence for Learning from Noisy {NLP} Datasets",
	author = "Bejan, Irina  and
	Sokolov, Artem  and
	Filippova, Katja",
	editor = "Bouamor, Houda  and
	Pino, Juan  and
	Bali, Kalika",
	booktitle = "Proceedings of the 2023 Conference on Empirical Methods in Natural Language Processing",
	month = dec,
	year = "2023",
	address = "Singapore",
	publisher = "Association for Computational Linguistics",
	url = "https://aclanthology.org/2023.emnlp-main.625/",
	doi = "10.18653/v1/2023.emnlp-main.625",
	pages = "10107--10121"
}

@article{chen2021evaluating,
	title={Evaluating large language models trained on code},
	author={Chen, Mark and Tworek, Jerry and Jun, Heewoo and Yuan, Qiming and Pinto, Henrique Ponde De Oliveira and Kaplan, Jared and Edwards, Harri and Burda, Yuri and Joseph, Nicholas and Brockman, Greg and others},
	journal={arXiv preprint arXiv:2107.03374},
	year={2021}
}

@article{jiang2023survey,
	author = {Jiang, Juyong and Wang, Fan and Shen, Jiasi and Kim, Sungju and Kim, Sunghun},
	title = {A Survey on Large Language Models for Code Generation},
	year = {2026},
	issue_date = {February 2026},
	publisher = {Association for Computing Machinery},
	address = {New York, NY, USA},
	volume = {35},
	number = {2},
	issn = {1049-331X},
	url = {https://doi.org/10.1145/3747588},
	doi = {10.1145/3747588},
	journal = {ACM Trans. Softw. Eng. Methodol.},
	month = jan,
	articleno = {58},
	numpages = {72}
}

@article{lialin2023scaling,
	title={Scaling down to scale up: A guide to parameter-efficient fine-tuning},
	author={Lialin, Vladislav and Deshpande, Vijeta and Rumshisky, Anna},
	journal={arXiv preprint arXiv:2303.15647},
	year={2023}
}

@inproceedings{zhou2023lima,
	author = {Zhou, Chunting and Liu, Pengfei and Xu, Puxin and Iyer, Srinivasan and Sun, Jiao and Mao, Yuning and Ma, Xuezhe and Efrat, Avia and Yu, Ping and YU, LILI and Zhang, Susan and Ghosh, Gargi and Lewis, Mike and Zettlemoyer, Luke and Levy, Omer},
	booktitle = {Advances in Neural Information Processing Systems},
	editor = {A. Oh and T. Naumann and A. Globerson and K. Saenko and M. Hardt and S. Levine},
	pages = {55006--55021},
	publisher = {Curran Associates, Inc.},
	title = {LIMA: Less Is More for Alignment},
	url = {https://proceedings.neurips.cc/paper_files/paper/2023/file/ac662d74829e4407ce1d126477f4a03a-Paper-Conference.pdf},
	volume = {36},
	year = {2023}
}

@article{chen2023maybe,
	title={Maybe only 0.5\% data is needed: A preliminary exploration of low training data instruction tuning},
	author={Chen, Hao and Zhang, Yiming and Zhang, Qi and Yang, Hantao and Hu, Xiaomeng and Ma, Xuetao and Yanggong, Yifan and Zhao, Junbo},
	journal={arXiv preprint arXiv:2305.09246},
	year={2023}
}

@inproceedings{
	lu2024instag,
	title={\#InsTag: Instruction Tagging for Analyzing Supervised Fine-tuning of Large Language Models},
	author={Keming Lu and Hongyi Yuan and Zheng Yuan and Runji Lin and Junyang Lin and Chuanqi Tan and Chang Zhou and Jingren Zhou},
	booktitle={The Twelfth International Conference on Learning Representations},
	year={2024},
	url={https://openreview.net/forum?id=pszewhybU9}
}

@article{li2025machine,
	author={Li, Na and Zhou, Chunyi and Gao, Yansong and Chen, Hui and Zhang, Zhi and Kuang, Boyu and Fu, Anmin},
	journal={IEEE Transactions on Neural Networks and Learning Systems}, 
	title={Machine Unlearning: Taxonomy, Metrics, Applications, Challenges, and Prospects}, 
	year={2025},
	volume={36},
	number={8},
	pages={13709-13729},
	doi={10.1109/TNNLS.2025.3530988}}

@inproceedings{chen2023unlearning,
	title = "Unlearning Bias in Language Models by Partitioning Gradients",
	author = "Yu, Charles  and
	Jeoung, Sullam  and
	Kasi, Anish  and
	Yu, Pengfei  and
	Ji, Heng",
	editor = "Rogers, Anna  and
	Boyd-Graber, Jordan  and
	Okazaki, Naoaki",
	booktitle = "Findings of the Association for Computational Linguistics: ACL 2023",
	month = jul,
	year = "2023",
	address = "Toronto, Canada",
	publisher = "Association for Computational Linguistics",
	url = "https://aclanthology.org/2023.findings-acl.375/",
	doi = "10.18653/v1/2023.findings-acl.375",
	pages = "6032--6048"
}

@article{wu2023self,
	title={Self-evolved diverse data sampling for efficient instruction tuning},
	author={Wu, Shengguang and Lu, Keming and Xu, Benfeng and Lin, Junyang and Su, Qi and Zhou, Chang},
	journal={arXiv preprint arXiv:2311.08182},
	year={2023}
}

@article{eldan2023whos,
	title={Who’s harry potter? approximate unlearning for LLMs},
	author={Eldan, Ronen and Russinovich, Mark},
	year={2023}
}

@inproceedings{jin2024rwku,
	author = {jin, Zhuoran and Cao, Pengfei and Wang, Chenhao and He, Zhitao and Yuan, Hongbang and Li, Jiachun and Chen, Yubo and Liu, Kang and Zhao, Jun},
	booktitle = {Advances in Neural Information Processing Systems},
	doi = {10.52202/079017-3117},
	editor = {A. Globerson and L. Mackey and D. Belgrave and A. Fan and U. Paquet and J. Tomczak and C. Zhang},
	pages = {98213--98263},
	publisher = {Curran Associates, Inc.},
	title = {RWKU: Benchmarking Real-World Knowledge Unlearning for Large Language Models},
	url = {https://proceedings.neurips.cc/paper_files/paper/2024/file/b1f78dfc9ca0156498241012aec4efa0-Paper-Datasets_and_Benchmarks_Track.pdf},
	volume = {37},
	year = {2024}
}

@inproceedings{de2021editing,
	title = "Editing Factual Knowledge in Language Models",
	author = "De Cao, Nicola  and
	Aziz, Wilker  and
	Titov, Ivan",
	editor = "Moens, Marie-Francine  and
	Huang, Xuanjing  and
	Specia, Lucia  and
	Yih, Scott Wen-tau",
	booktitle = "Proceedings of the 2021 Conference on Empirical Methods in Natural Language Processing",
	month = nov,
	year = "2021",
	address = "Online and Punta Cana, Dominican Republic",
	publisher = "Association for Computational Linguistics",
	url = "https://aclanthology.org/2021.emnlp-main.522/",
	doi = "10.18653/v1/2021.emnlp-main.522",
	pages = "6491--6506"
}

@inproceedings{jang2022knowledge,
	title = "Knowledge Unlearning for Mitigating Privacy Risks in Language Models",
	author = "Jang, Joel  and
	Yoon, Dongkeun  and
	Yang, Sohee  and
	Cha, Sungmin  and
	Lee, Moontae  and
	Logeswaran, Lajanugen  and
	Seo, Minjoon",
	editor = "Rogers, Anna  and
	Boyd-Graber, Jordan  and
	Okazaki, Naoaki",
	booktitle = "Proceedings of the 61st Annual Meeting of the Association for Computational Linguistics (Volume 1: Long Papers)",
	month = jul,
	year = "2023",
	address = "Toronto, Canada",
	publisher = "Association for Computational Linguistics",
	url = "https://aclanthology.org/2023.acl-long.805/",
	doi = "10.18653/v1/2023.acl-long.805",
	pages = "14389--14408"
}

@article{maini2024tofu,
	title={Tofu: A task of fictitious unlearning for llms},
	author={Maini, Pratyush and Feng, Zhili and Schwarzschild, Avi and Lipton, Zachary C and Kolter, J Zico},
	journal={arXiv preprint arXiv:2401.06121},
	year={2024}
}

@inproceedings{wang2025rethinking,
	title={Rethinking LLM Unlearning Objectives: A Gradient Perspective and Go Beyond}, 
	author={Qizhou Wang and Jin Peng Zhou and Zhanke Zhou and Saebyeol Shin and Bo Han and Kilian Q Weinberger},
	booktitle = {International Conference on Learning Representations},
	year = {2025}
}

@inproceedings{lizzo2024unlearn,
	title = "{UNLEARN} Efficient Removal of Knowledge in Large Language Models",
	author = "Lizzo, Tyler  and
	Heck, Larry",
	editor = "Chiruzzo, Luis  and
	Ritter, Alan  and
	Wang, Lu",
	booktitle = "Findings of the Association for Computational Linguistics: NAACL 2025",
	month = apr,
	year = "2025",
	address = "Albuquerque, New Mexico",
	publisher = "Association for Computational Linguistics",
	url = "https://aclanthology.org/2025.findings-naacl.405/",
	doi = "10.18653/v1/2025.findings-naacl.405",
	pages = "7272--7283",
	ISBN = "979-8-89176-195-7"
}

@article{zhang2024negative,
	title={Negative preference optimization: From catastrophic collapse to effective unlearning},
	author={Zhang, Ruiqi and Lin, Licong and Bai, Yu and Mei, Song},
	journal={arXiv preprint arXiv:2404.05868},
	year={2024}
}

@inproceedings{wang2024towards,
	title={Towards Effective Evaluations and Comparison for LLM Unlearning Methods}, 
	author={Qizhou Wang and Bo Han and Puning Yang and Jianing Zhu and Tongliang Liu and Masashi Sugiyama},
	booktitle = {International Conference on Learning Representations},
	year = {2025}
}

@inproceedings{yao2024large,
	author = {Yao, Yuanshun and Xu, Xiaojun and YangLiu},
	booktitle = {Advances in Neural Information Processing Systems},
	doi = {10.52202/079017-3346},
	editor = {A. Globerson and L. Mackey and D. Belgrave and A. Fan and U. Paquet and J. Tomczak and C. Zhang},
	pages = {105425--105475},
	publisher = {Curran Associates, Inc.},
	title = {Large Language Model Unlearning},
	url = {https://proceedings.neurips.cc/paper_files/paper/2024/file/be52acf6bccf4a8c0a90fe2f5cfcead3-Paper-Conference.pdf},
	volume = {37},
	year = {2024}
}

@inproceedings{yao2024machine,
	title = "Machine Unlearning of Pre-trained Large Language Models",
	author = "Yao, Jin  and
	Chien, Eli  and
	Du, Minxin  and
	Niu, Xinyao  and
	Wang, Tianhao  and
	Cheng, Zezhou  and
	Yue, Xiang",
	editor = "Ku, Lun-Wei  and
	Martins, Andre  and
	Srikumar, Vivek",
	booktitle = "Proceedings of the 62nd Annual Meeting of the Association for Computational Linguistics (Volume 1: Long Papers)",
	month = aug,
	year = "2024",
	address = "Bangkok, Thailand",
	publisher = "Association for Computational Linguistics",
	url = "https://aclanthology.org/2024.acl-long.457/",
	doi = "10.18653/v1/2024.acl-long.457",
	pages = "8403--8419"
}

@inproceedings{tian2024forget,
	title = "To Forget or Not? Towards Practical Knowledge Unlearning for Large Language Models",
	author = "Tian, Bozhong  and
	Liang, Xiaozhuan  and
	Cheng, Siyuan  and
	Liu, Qingbin  and
	Wang, Mengru  and
	Sui, Dianbo  and
	Chen, Xi  and
	Chen, Huajun  and
	Zhang, Ningyu",
	editor = "Al-Onaizan, Yaser  and
	Bansal, Mohit  and
	Chen, Yun-Nung",
	booktitle = "Findings of the Association for Computational Linguistics: EMNLP 2024",
	month = nov,
	year = "2024",
	address = "Miami, Florida, USA",
	publisher = "Association for Computational Linguistics",
	url = "https://aclanthology.org/2024.findings-emnlp.82/",
	doi = "10.18653/v1/2024.findings-emnlp.82",
	pages = "1524--1537"
}

@inproceedings{
	cha2024towards,
	title={Towards Robust and Cost-Efficient Knowledge Unlearning for Large Language Models},
	author={Sungmin Cha and Sungjun Cho and Dasol Hwang and Moontae Lee},
	booktitle={Adaptive Foundation Models: Evolving AI for Personalized and Efficient Learning},
	year={2024},
	url={https://openreview.net/forum?id=1AWLjICe5P}
}

@inproceedings{xia2024less,
	author = {Xia, Mengzhou and Malladi, Sadhika and Gururangan, Suchin and Arora, Sanjeev and Chen, Danqi},
	title = {LESS: selecting influential data for targeted instruction tuning},
	year = {2024},
	publisher = {JMLR.org},
	booktitle = {Proceedings of the 41st International Conference on Machine Learning},
	articleno = {2221},
	numpages = {29},
	location = {Vienna, Austria},
	series = {ICML'24}
}

@InProceedings{koh2017understanding,
	title = 	 {Understanding Black-box Predictions via Influence Functions},
	author =       {Pang Wei Koh and Percy Liang},
	booktitle = 	 {Proceedings of the 34th International Conference on Machine Learning},
	pages = 	 {1885--1894},
	year = 	 {2017},
	editor = 	 {Precup, Doina and Teh, Yee Whye},
	volume = 	 {70},
	series = 	 {Proceedings of Machine Learning Research},
	month = 	 {06--11 Aug},
	publisher =    {PMLR},
	url = 	 {https://proceedings.mlr.press/v70/koh17a.html}
}

@inproceedings{
	pareja2024unveiling,
	title={Unveiling the Secret Recipe: A Guide For Supervised Fine-Tuning Small {LLM}s},
	author={Aldo Pareja and Nikhil Shivakumar Nayak and Hao Wang and Krishnateja Killamsetty and Shivchander Sudalairaj and Wenlong Zhao and Seungwook Han and Abhishek Bhandwaldar and Guangxuan Xu and Kai Xu and Ligong Han and Luke Inglis and Akash Srivastava},
	booktitle={The Thirteenth International Conference on Learning Representations},
	year={2025},
	url={https://openreview.net/forum?id=eENHKMTOfW}
}

@InProceedings{zhou2023dataset,
	author    = {Zhou, Daquan and Wang, Kai and Gu, Jianyang and Peng, Xiangyu and Lian, Dongze and Zhang, Yifan and You, Yang and Feng, Jiashi},
	title     = {Dataset Quantization},
	booktitle = {Proceedings of the IEEE/CVF International Conference on Computer Vision (ICCV)},
	month     = {October},
	year      = {2023},
	pages     = {17205-17216}
}

@article{luo2024robustft,
	title={Robustft: Robust supervised fine-tuning for large language models under noisy response},
	author={Luo, Junyu and Luo, Xiao and Ding, Kaize and Yuan, Jingyang and Xiao, Zhiping and Zhang, Ming},
	journal={arXiv preprint arXiv:2412.14922},
	year={2024}
}

@inproceedings{
	zhang2025best,
	title={The Best Instruction-Tuning Data are Those That Fit},
	author={Dylan Zhang and Qirun Dai and Hao Peng},
	booktitle={The Thirty-ninth Annual Conference on Neural Information Processing Systems},
	year={2025},
	url={https://openreview.net/forum?id=4jFSekBaDT}
}

@inproceedings{dai2025improving,
	title = "Improving Influence-based Instruction Tuning Data Selection for Balanced Learning of Diverse Capabilities",
	author = "Dai, Qirun  and
	Zhang, Dylan  and
	Ma, Jiaqi W.  and
	Peng, Hao",
	editor = "Christodoulopoulos, Christos  and
	Chakraborty, Tanmoy  and
	Rose, Carolyn  and
	Peng, Violet",
	booktitle = "Findings of the Association for Computational Linguistics: EMNLP 2025",
	month = nov,
	year = "2025",
	address = "Suzhou, China",
	publisher = "Association for Computational Linguistics",
	url = "https://aclanthology.org/2025.findings-emnlp.373/",
	doi = "10.18653/v1/2025.findings-emnlp.373",
	pages = "7079--7102",
	ISBN = "979-8-89176-335-7"
}

@misc{databricks2023dolly,
	author    = {Mike Conover and Matt Hayes and Ankit Mathur and Jianwei Xie and Jun Wan and Sam Shah and Ali Ghodsi and Patrick Wendell and Matei Zaharia and Reynold Xin},
	title     = {Free Dolly: Introducing the World's First Truly Open Instruction-Tuned LLM},
	year      = {2023},
	url       = {https://www.databricks.com/blog/2023/04/12/dolly-first-open-commercially-viable-instruction-tuned-llm},
	urldate   = {2023-06-30}
}

@inproceedings{openassistant2023,
	author = {K\"{o}pf, Andreas and Kilcher, Yannic and von R\"{u}tte, Dimitri and Anagnostidis, Sotiris and Tam, Zhi Rui and Stevens, Keith and Barhoum, Abdullah and Nguyen, Duc and Stanley, Oliver and Nagyfi, Rich\'{a}rd and ES, Shahul and Suri, Sameer and Glushkov, David and Dantuluri, Arnav and Maguire, Andrew and Schuhmann, Christoph and Nguyen, Huu and Mattick, Alexander},
	booktitle = {Advances in Neural Information Processing Systems},
	editor = {A. Oh and T. Naumann and A. Globerson and K. Saenko and M. Hardt and S. Levine},
	pages = {47669--47681},
	publisher = {Curran Associates, Inc.},
	title = {OpenAssistant Conversations - Democratizing Large Language Model Alignment},
	url = {https://proceedings.neurips.cc/paper_files/paper/2023/file/949f0f8f32267d297c2d4e3ee10a2e7e-Paper-Datasets_and_Benchmarks.pdf},
	volume = {36},
	year = {2023}
}

@misc{taori2023alpaca,
	author = {Rohan Taori and Ishaan Gulrajani and Tianyi Zhang and Yann Dubois and Xuechen Li and Carlos Guestrin and Percy Liang and Tatsunori B. Hashimoto },
	title = {Stanford Alpaca: An Instruction-following LLaMA model},
	year = {2023},
	publisher = {GitHub},
	journal = {GitHub repository},
	howpublished = {\url{https://github.com/tatsu-lab/stanford_alpaca}},
}

@article{xu2023wizardlm,
	title={Wizardlm: Empowering large language models to follow complex instructions},
	author={Xu, Can and Sun, Qingfeng and Zheng, Kai and Geng, Xiubo and Zhao, Pu and Feng, Jiazhan and Tao, Chongyang and Jiang, Daxin},
	journal={arXiv preprint arXiv:2304.12244},
	year={2023}
}

@inproceedings{lin2022truthfulqa,
	title = "{T}ruthful{QA}: Measuring How Models Mimic Human Falsehoods",
	author = "Lin, Stephanie  and
	Hilton, Jacob  and
	Evans, Owain",
	editor = "Muresan, Smaranda  and
	Nakov, Preslav  and
	Villavicencio, Aline",
	booktitle = "Proceedings of the 60th Annual Meeting of the Association for Computational Linguistics (Volume 1: Long Papers)",
	month = may,
	year = "2022",
	address = "Dublin, Ireland",
	publisher = "Association for Computational Linguistics",
	url = "https://aclanthology.org/2022.acl-long.229/",
	doi = "10.18653/v1/2022.acl-long.229",
	pages = "3214--3252"
}

@inproceedings{clark2019boolq,
	title = "{B}ool{Q}: Exploring the Surprising Difficulty of Natural Yes/No Questions",
	author = "Clark, Christopher  and
	Lee, Kenton  and
	Chang, Ming-Wei  and
	Kwiatkowski, Tom  and
	Collins, Michael  and
	Toutanova, Kristina",
	editor = "Burstein, Jill  and
	Doran, Christy  and
	Solorio, Thamar",
	booktitle = "Proceedings of the 2019 Conference of the North {A}merican Chapter of the Association for Computational Linguistics: Human Language Technologies, Volume 1 (Long and Short Papers)",
	month = jun,
	year = "2019",
	address = "Minneapolis, Minnesota",
	publisher = "Association for Computational Linguistics",
	url = "https://aclanthology.org/N19-1300/",
	doi = "10.18653/v1/N19-1300",
	pages = "2924--2936"
}

@inproceedings{liu2020logiqa,
	title     = {LogiQA: A Challenge Dataset for Machine Reading Comprehension with Logical Reasoning},
	author    = {Liu, Jian and Cui, Leyang and Liu, Hanmeng and Huang, Dandan and Wang, Yile and Zhang, Yue},
	booktitle = {Proceedings of the Twenty-Ninth International Joint Conference on
	Artificial Intelligence, {IJCAI-20}},
	publisher = {International Joint Conferences on Artificial Intelligence Organization},
	editor    = {Christian Bessiere},
	pages     = {3622--3628},
	year      = {2020},
	month     = {7},
	note      = {Main track},
	doi       = {10.24963/ijcai.2020/501},
	url       = {https://doi.org/10.24963/ijcai.2020/501},
}

@article{clark2020tydiqa,
	author = {Clark, Jonathan H. and Choi, Eunsol and Collins, Michael and Garrette, Dan and Kwiatkowski, Tom and Nikolaev, Vitaly and Palomaki, Jennimaria},
	title = {TyDi QA: A Benchmark for Information-Seeking Question Answering in Typologically Diverse Languages},
	journal = {Transactions of the Association for Computational Linguistics},
	volume = {8},
	pages = {454-470},
	year = {2020},
	month = {07},
	issn = {2307-387X},
	doi = {10.1162/tacl_a_00317},
	url = {https://doi.org/10.1162/tacl_a_00317},
	eprint = {https://direct.mit.edu/tacl/article-pdf/doi/10.1162/tacl_a_00317/1923348/tacl_a_00317.pdf},
}

@inproceedings{miao2021diverse,
	title = "A Diverse Corpus for Evaluating and Developing {E}nglish Math Word Problem Solvers",
	author = "Miao, Shen-yun  and
	Liang, Chao-Chun  and
	Su, Keh-Yih",
	editor = "Jurafsky, Dan  and
	Chai, Joyce  and
	Schluter, Natalie  and
	Tetreault, Joel",
	booktitle = "Proceedings of the 58th Annual Meeting of the Association for Computational Linguistics",
	month = jul,
	year = "2020",
	address = "Online",
	publisher = "Association for Computational Linguistics",
	url = "https://aclanthology.org/2020.acl-main.92/",
	doi = "10.18653/v1/2020.acl-main.92",
	pages = "975--984"
}

@article{dubey2024llama3,
	title={The llama 3 herd of models},
	author={Grattafiori, Aaron and Dubey, Abhimanyu and Jauhri, Abhinav and Pandey, Abhinav and Kadian, Abhishek and Al-Dahle, Ahmad and Letman, Aiesha and Mathur, Akhil and Schelten, Alan and Vaughan, Alex and others},
	journal={arXiv preprint arXiv:2407.21783},
	year={2024}
}

@inproceedings{wang2023camels,
	author = {Wang, Yizhong and Ivison, Hamish and Dasigi, Pradeep and Hessel, Jack and Khot, Tushar and Chandu, Khyathi and Wadden, David and MacMillan, Kelsey and Smith, Noah A and Beltagy, Iz and Hajishirzi, Hannaneh},
	booktitle = {Advances in Neural Information Processing Systems},
	editor = {A. Oh and T. Naumann and A. Globerson and K. Saenko and M. Hardt and S. Levine},
	pages = {74764--74786},
	publisher = {Curran Associates, Inc.},
	title = {How Far Can Camels Go? Exploring the State of Instruction Tuning on Open Resources},
	url = {https://proceedings.neurips.cc/paper_files/paper/2023/file/ec6413875e4ab08d7bc4d8e225263398-Paper-Datasets_and_Benchmarks.pdf},
	volume = {36},
	year = {2023}
}

@inproceedings{
	hu2022lora,
	title={Lo{RA}: Low-Rank Adaptation of Large Language Models},
	author={Edward J Hu and Yelong Shen and Phillip Wallis and Zeyuan Allen-Zhu and Yuanzhi Li and Shean Wang and Lu Wang and Weizhu Chen},
	booktitle={International Conference on Learning Representations},
	year={2022},
	url={https://openreview.net/forum?id=nZeVKeeFYf9}
}

@inproceedings{
	loshchilov2017decoupled,
	title={Decoupled Weight Decay Regularization},
	author={Ilya Loshchilov and Frank Hutter},
	booktitle={International Conference on Learning Representations},
	year={2019},
	url={https://openreview.net/forum?id=Bkg6RiCqY7},
}

@article{schulman2017proximal,
	title={Proximal policy optimization algorithms},
	author={Schulman, John and Wolski, Filip and Dhariwal, Prafulla and Radford, Alec and Klimov, Oleg},
	journal={arXiv preprint arXiv:1707.06347},
	year={2017}
}

@inproceedings{rafailov2023direct,
	author = {Rafailov, Rafael and Sharma, Archit and Mitchell, Eric and Manning, Christopher D and Ermon, Stefano and Finn, Chelsea},
	booktitle = {Advances in Neural Information Processing Systems},
	editor = {A. Oh and T. Naumann and A. Globerson and K. Saenko and M. Hardt and S. Levine},
	pages = {53728--53741},
	publisher = {Curran Associates, Inc.},
	title = {Direct Preference Optimization: Your Language Model is Secretly a Reward Model},
	url = {https://proceedings.neurips.cc/paper_files/paper/2023/file/a85b405ed65c6477a4fe8302b5e06ce7-Paper-Conference.pdf},
	volume = {36},
	year = {2023}
}

@InProceedings{ethayarajh2024kto,
	title = 	 {Model Alignment as Prospect Theoretic Optimization},
	author =       {Ethayarajh, Kawin and Xu, Winnie and Muennighoff, Niklas and Jurafsky, Dan and Kiela, Douwe},
	booktitle = 	 {Proceedings of the 41st International Conference on Machine Learning},
	pages = 	 {12634--12651},
	year = 	 {2024},
	editor = 	 {Salakhutdinov, Ruslan and Kolter, Zico and Heller, Katherine and Weller, Adrian and Oliver, Nuria and Scarlett, Jonathan and Berkenkamp, Felix},
	volume = 	 {235},
	series = 	 {Proceedings of Machine Learning Research},
	month = 	 {21--27 Jul},
	publisher =    {PMLR},
	url = 	 {https://proceedings.mlr.press/v235/ethayarajh24a.html}
}

@InProceedings{azar2023general,
	title = 	 {A General Theoretical Paradigm to Understand Learning from Human Preferences},
	author =       {Gheshlaghi Azar, Mohammad and Daniel Guo, Zhaohan and Piot, Bilal and Munos, Remi and Rowland, Mark and Valko, Michal and Calandriello, Daniele},
	booktitle = 	 {Proceedings of The 27th International Conference on Artificial Intelligence and Statistics},
	pages = 	 {4447--4455},
	year = 	 {2024},
	editor = 	 {Dasgupta, Sanjoy and Mandt, Stephan and Li, Yingzhen},
	volume = 	 {238},
	series = 	 {Proceedings of Machine Learning Research},
	month = 	 {02--04 May},
	publisher =    {PMLR},
	url = 	 {https://proceedings.mlr.press/v238/gheshlaghi-azar24a.html}
}

@inproceedings{xu2024contrastive,
	author = {Xu, Haoran and Sharaf, Amr and Chen, Yunmo and Tan, Weiting and Shen, Lingfeng and Van Durme, Benjamin and Murray, Kenton and Kim, Young Jin},
	title = {Contrastive preference optimization: pushing the boundaries of LLM performance in machine translation},
	year = {2024},
	publisher = {JMLR.org},
	booktitle = {Proceedings of the 41st International Conference on Machine Learning},
	articleno = {2275},
	numpages = {21},
	location = {Vienna, Austria},
	series = {ICML'24}
}

@inproceedings{hong2024orpo,
	title = "{ORPO}: Monolithic Preference Optimization without Reference Model",
	author = "Hong, Jiwoo  and
	Lee, Noah  and
	Thorne, James",
	editor = "Al-Onaizan, Yaser  and
	Bansal, Mohit  and
	Chen, Yun-Nung",
	booktitle = "Proceedings of the 2024 Conference on Empirical Methods in Natural Language Processing",
	month = nov,
	year = "2024",
	address = "Miami, Florida, USA",
	publisher = "Association for Computational Linguistics",
	url = "https://aclanthology.org/2024.emnlp-main.626/",
	doi = "10.18653/v1/2024.emnlp-main.626",
	pages = "11170--11189"
}

@inproceedings{meng2024simpo,
	author = {Meng, Yu and Xia, Mengzhou and Chen, Danqi},
	booktitle = {Advances in Neural Information Processing Systems},
	doi = {10.52202/079017-3946},
	editor = {A. Globerson and L. Mackey and D. Belgrave and A. Fan and U. Paquet and J. Tomczak and C. Zhang},
	pages = {124198--124235},
	publisher = {Curran Associates, Inc.},
	title = {SimPO: Simple Preference Optimization with a Reference-Free Reward},
	url = {https://proceedings.neurips.cc/paper_files/paper/2024/file/e099c1c9699814af0be873a175361713-Paper-Conference.pdf},
	volume = {37},
	year = {2024}
}

@article{touvron2023llama,
	title={Llama 2: Open foundation and fine-tuned chat models},
	author={Touvron, Hugo and Martin, Louis and Stone, Kevin and Albert, Peter and Almahairi, Amjad and Babaei, Yasmine and Bashlykov, Nikolay and Batra, Soumya and Bhargava, Prajjwal and Bhosale, Shruti and others},
	journal={arXiv preprint arXiv:2307.09288},
	year={2023}
}

@InProceedings{zeng2024token,
	title = 	 {Token-level Direct Preference Optimization},
	author =       {Zeng, Yongcheng and Liu, Guoqing and Ma, Weiyu and Yang, Ning and Zhang, Haifeng and Wang, Jun},
	booktitle = 	 {Proceedings of the 41st International Conference on Machine Learning},
	pages = 	 {58348--58365},
	year = 	 {2024},
	editor = 	 {Salakhutdinov, Ruslan and Kolter, Zico and Heller, Katherine and Weller, Adrian and Oliver, Nuria and Scarlett, Jonathan and Berkenkamp, Felix},
	volume = 	 {235},
	series = 	 {Proceedings of Machine Learning Research},
	month = 	 {21--27 Jul},
	publisher =    {PMLR},
	url = 	 {https://proceedings.mlr.press/v235/zeng24c.html}
}

@article{radford2019language,
	author = {Radford, Alec and Wu, Jeffrey and Child, Rewon and Luan, David and Amodei, Dario and Sutskever, Ilya},
	journal = {OpenAI},
	note = {Accessed: 2024-11-15},
	title = {Language Models are Unsupervised Multitask Learners},
	url = {https://cdn.openai.com/better-language-models/language_models_are_unsupervised_multitask_learners.pdf},
	year = 2019
}

@misc{yang2024qwen25,
	title={Qwen2.5 Technical Report}, 
	author={Qwen and : and An Yang and Baosong Yang and Beichen Zhang and Binyuan Hui and Bo Zheng and Bowen Yu and Chengyuan Li and Dayiheng Liu and Fei Huang and Haoran Wei and Huan Lin and Jian Yang and Jianhong Tu and Jianwei Zhang and Jianxin Yang and Jiaxi Yang and Jingren Zhou and Junyang Lin and Kai Dang and Keming Lu and Keqin Bao and Kexin Yang and Le Yu and Mei Li and Mingfeng Xue and Pei Zhang and Qin Zhu and Rui Men and Runji Lin and Tianhao Li and Tianyi Tang and Tingyu Xia and Xingzhang Ren and Xuancheng Ren and Yang Fan and Yang Su and Yichang Zhang and Yu Wan and Yuqiong Liu and Zeyu Cui and Zhenru Zhang and Zihan Qiu},
	year={2025},
	eprint={2412.15115},
	archivePrefix={arXiv},
	primaryClass={cs.CL},
	url={https://arxiv.org/abs/2412.15115}, 
}

@article{black2021gptneo,
	title={Gpt-neo: Large scale autoregressive language modeling with mesh-tensorflow},
	author={Black, Sid and Leo, Gao and Wang, Phil and Leahy, Connor and Biderman, Stella},
	journal={Zenodo},
	year={2021}
}

@article{zhou2023ifeval,
	title={Instruction-following evaluation for large language models},
	author={Zhou, Jeffrey and Lu, Tianjian and Mishra, Swaroop and Brahma, Siddhartha and Basu, Sujoy and Luan, Yi and Zhou, Denny and Hou, Le},
	journal={arXiv preprint arXiv:2311.07911},
	year={2023}
}

@article{clark2018arc,
	title={Think you have solved question answering? try arc, the ai2 reasoning challenge},
	author={Clark, Peter and Cowhey, Isaac and Etzioni, Oren and Khot, Tushar and Sabharwal, Ashish and Schoenick, Carissa and Tafjord, Oyvind},
	journal={arXiv preprint arXiv:1803.05457},
	year={2018}
}

@inproceedings{paperno2016lambada,
	title = "The {LAMBADA} dataset: Word prediction requiring a broad discourse context",
	author = "Paperno, Denis  and
	Kruszewski, Germ{\'a}n  and
	Lazaridou, Angeliki  and
	Pham, Ngoc Quan  and
	Bernardi, Raffaella  and
	Pezzelle, Sandro  and
	Baroni, Marco  and
	Boleda, Gemma  and
	Fern{\'a}ndez, Raquel",
	editor = "Erk, Katrin  and
	Smith, Noah A.",
	booktitle = "Proceedings of the 54th Annual Meeting of the Association for Computational Linguistics (Volume 1: Long Papers)",
	month = aug,
	year = "2016",
	address = "Berlin, Germany",
	publisher = "Association for Computational Linguistics",
	url = "https://aclanthology.org/P16-1144/",
	doi = "10.18653/v1/P16-1144",
	pages = "1525--1534"
}

\clearpage
\appendix
\section{Datasets details}
\label{app:dataset}
Table \ref{datasets-table} provides comprehensive information about the datasets used to create training corpus, including their quality assessment, size, total length of samples, and source.

\begin{table}[ht]
  \caption{Datasets attributes}
  \label{datasets-table}
  \centering
  \begin{tabular}{p{2.5cm}p{1.8cm}p{1.6cm}p{1.5cm}p{2.8cm}}
    \toprule
    Dataset & Data Quality & Size & Length & Resource \\
    \midrule
    Dolly & High & 15.01k & Varied & Human-annotated \\
    Flan\_v2 & High & 100k & Varied & Human-annotated \\
    Open Assistant 1 & Moderate & 33.92k & Varied & Human-annotated \\
    Stanford Alpaca & High & 52k & Varied & LLM-generated \\
    WizardLM & High & 100k & Longer & LLM-generated \\
    \bottomrule
  \end{tabular}
\end{table}

\section{Additional experimental results}
\label{app:additional-results}

\subsection{LLaMA-2-13B results}
\label{app:llama2-13b}

To further validate the robustness and scalability of our forgetting mechanism, we conducted additional experiments using LLaMA-2-13B as the base model. These results provide additional evidence that our approach consistently improves performance across different model architectures and scales, extending beyond the LLaMA-3.x series reported in the main paper.

The results in Table~\ref{tab:llama2-13b-results} demonstrate that our forgetting method maintains its effectiveness with larger models, achieving a 6.16\% improvement over standard SFT and a 4.16\% improvement over the ignoring baseline. This consistency across model scales (from 1B to 13B parameters) reinforces the generalizability of our approach and suggests that the forgetting mechanism provides fundamental benefits for supervised fine-tuning regardless of model size or architecture.

\begin{table}[ht]
\caption{Performance comparison of different methods across five benchmarks using LLaMA-2-13B as the base model. Results show accuracy (\%) for TruthfulQA, BoolQ, LogiQA, and ASDiv, and one-shot F1 score for TydiQA. Bold values demonstrate best performance on each benchmark. Our proposed Forgetting method achieves significant improvements across different benchmarks, with an average improvement of 6.16\% over standard SFT and 4.16\% over the ignoring approach.}
\label{tab:llama2-13b-results}
\centering
\begin{tabular}{lcccccc}
\toprule[1.5pt]
Method & \multicolumn{6}{c}{Dataset} \\
\cmidrule(lr){2-7}
 & TruthfulQA & BoolQ & LogiQA & TydiQA & ASDiV & AVG \\
\midrule[1.5pt]
\multicolumn{7}{c}{Base model: LLaMA-2-13B} \\
\midrule[1.5pt]
Base & 36.73 & 80.67 & 26.05 & 34.27 & 0.35 & 35.61 \\
Full Tokens (standard SFT) & 42.65 & 82.24 & 27.44 & 36.77 & 8.76 & 39.57 \\
Ignoring & 43.01 & \textbf{84.50} & 27.29 & 38.39 & 15.34 & 41.71 \\
\midrule
Forgetting (Ours) & \textbf{52.82} & 84.13 & \textbf{27.95} & \textbf{48.71} & \textbf{17.80} & \textbf{46.28} \\
\bottomrule[1.5pt]
\end{tabular}
\end{table}

\subsection{Impact of Reference Dataset Duplicates}
\label{app:reference-duplicates}

We conducted additional experiments to investigate the robustness of our approach when the reference dataset contains duplicate samples. However our pipeline's preprocessing step removes duplicate samples from the both training and references datasets, this analysis is important for understanding how data quality in the reference model training affects the overall forgetting mechanism performance.

Table~\ref{tab:duplicate-reference-results} shows results using LLaMA-3.2-3B when the reference dataset includes duplicate samples. Interestingly, our forgetting method remains effective even under these suboptimal reference conditions, achieving a 4.93\% improvement over standard SFT and a 2.05\% improvement over the ignoring baseline. This demonstrates the robustness of our influence-based token quality assessment even when the reference model is trained on imperfect data, suggesting that our approach can handle practical scenarios where perfect data curation is not feasible.

\begin{table}[ht]
\caption{Performance comparison with duplicate samples in reference dataset using LLaMA-3.2-3B as base model. Results show mean values with standard deviations from 3 independent training runs. Our forgetting method maintains effectiveness even with imperfect reference data quality.}
\label{tab:duplicate-reference-results}
\centering
\resizebox{\textwidth}{!}{%
\begin{tabular}{lcccccc}
\toprule[1.5pt]
Method & \multicolumn{6}{c}{Dataset} \\
\cmidrule(lr){2-7}
 & TruthfulQA & BoolQ & LogiQA & TydiQA & ASDiV & AVG \\
\midrule[1.5pt]
\multicolumn{7}{c}{Base model: LLaMA-3.2-3B (Reference with Duplicates)} \\
\midrule[1.5pt]
Base & 39.45±0 & 73.04±0 & 22.17±0 & 21.12±0 & 31.24±0 & 37.40±0 \\
Full Tokens (standard SFT) & 42.95±0.47 & 72.54±0.59 & 25.51±0.21 & 44.04±0.27 & 49.46±0.14 & 46.90±0.16 \\
Ignoring & 49.91±0.39 & 75.60±0.86 & 24.99±0.35 & 48.61±0.20 & \textbf{49.81±0.01} & 49.78±0.22 \\
\midrule
Forgetting (Ours) & \textbf{51.09±0.54} & \textbf{77.00±0.09} & \textbf{26.57±0.08} & \textbf{54.88±0.29} & 49.60±0.14 & \textbf{51.83±0.11} \\
\bottomrule[1.5pt]
\end{tabular}
}
\end{table}

\subsection{Evaluation on Diverse Model Architectures}
\label{app:diverse-models}

To demonstrate the broad applicability of our forgetting mechanism, we extended our evaluation to additional model architectures beyond the LLaMA family. Specifically, we conducted experiments on Qwen2.5-3B and GPT-Neo-2.7B, evaluating performance across four diverse benchmarks including Instruction-Following (IFEval), ARC-Challenge, LAMBADA, and Arithmetic. The characteristics of these evaluation benchmarks are detailed in Table~\ref{tab:diverse-benchmarks}.

For these experiments, we maintained our LLaMA-3.2-3B experimental setup and hyperparameters, as described in Section~\ref{training-config}. The results presented in Table~\ref{tab:diverse-models-results} show that our forgetting method consistently outperforms both standard SFT (full tokens) and the ignoring baseline across both model architectures. On Qwen2.5-3B, our method achieves an average performance of 59.01\%, representing a 16.49\% improvement over standard SFT and a 5.33\% improvement over the ignoring approach. Similarly, on GPT-Neo-2.7B, our forgetting mechanism attains 28.15\% average performance, demonstrating a 4.37\% improvement over standard SFT and a 3.56\% improvement over ignoring. These results confirm that the effectiveness of our forgetting mechanism generalizes well across diverse model architectures and evaluation tasks, validating its broad applicability for improving SFT of large language models.

\begin{table}[ht]
  \caption{Characteristics of diverse evaluation benchmarks}
  \label{tab:diverse-benchmarks}
  \centering
  \begin{tabular}{llll}
    \toprule
    Dataset & Focus Area & Data Size & Question Length \\
    \midrule
    IFEval & Instruction Following & 541 & Varied \\
    ARC-Challenge & Scientific Reasoning & 1,172 & Medium \\
    LAMBADA & Language Modeling & 5,153 & Short \\
    Arithmetic & Mathematical Computation & Varied & Short \\
    \bottomrule
  \end{tabular}
\end{table}

\begin{table}[ht]
\caption{Performance comparison across diverse model architectures and benchmarks. Results show accuracy (\%) for all benchmarks. Bold values indicate best performance. Our forgetting method demonstrates consistent improvements across different model families (Qwen and GPT-Neo), validating its broad applicability beyond the LLaMA architecture family.}
\label{tab:diverse-models-results}
\centering
\renewcommand{\arraystretch}{1.3}
\begin{tabular}{llcccccc}
\toprule[1.5pt]
Model & Method & IFEval & ARC-Challenge & LAMBADA & Arithmetic & AVG \\
\midrule[1.5pt]
\multicolumn{7}{c}{Base model: Qwen2.5-3B} \\
\midrule[1.5pt]
& Base & 23.13 & 44.70 & 66.72 & 13.16 & 36.93 \\
Qwen2.5-3B & Full Tokens & 19.59 & 43.66 & 68.64 & 38.19 & 42.52 \\
& Ignoring & 19.22 & \textbf{45.63} & 68.82 & 81.03 & 53.68 \\
& Forgetting (Ours) & \textbf{33.49} & 45.39 & \textbf{69.76} & \textbf{87.40} & \textbf{59.01} \\
\midrule[1.5pt]
\multicolumn{7}{c}{Base model: GPT-Neo-2.7B} \\
\midrule[1.5pt]
& Base & 1.90 & 27.48 & 61.74 & 0.43 & 22.89 \\
GPT-Neo-2.7B & Full Tokens & 1.49 & 29.72 & 63.05 & 0.84 & 23.78 \\
& Ignoring & 0.19 & 30.55 & 66.15 & 1.46 & 24.59 \\
& Forgetting (Ours) & \textbf{11.66} & \textbf{31.63} & \textbf{67.29} & \textbf{2.03} & \textbf{28.15} \\
\bottomrule[1.5pt]
\end{tabular}
\end{table}

\subsection{Hyperparameter sensitivity analysis}
\label{app:hyperparameter-sensitivity}
Table~\ref{tab:hyperparameter-sensitivity} presents comprehensive results across different combinations of $t_{\min}$ and $t_{\max}$ values using LLaMA-3.2-3B. The results demonstrate remarkable stability, with performance variations remaining small across different hyperparameter settings (standard deviation < 0.5\% across configurations). This robustness ensures that our method maintains superiority over baselines without requiring extensive hyperparameter tuning. The stability is partly attributed to the inherent robustness of large language models and their extensive pre-trained knowledge, which provides a strong foundation that is resilient to moderate changes in fine-tuning parameters.

\begin{table}[ht]
\caption{Hyperparameter sensitivity analysis for $t_{\min}$ and $t_{\max}$ using LLaMA-3.2-3B with fixed $\rho=0.7$. Results demonstrate robustness across different parameter combinations.}
\label{tab:hyperparameter-sensitivity}
\centering
\small
\renewcommand{\arraystretch}{1.1}
\begin{tabular}{cccccccc}
\toprule
$t_{\min}$ & $t_{\max}$ & TruthfulQA & BoolQ & LogiQA & TydiQA & ASDiV & AVG \\
\midrule
0.00001 & 0.45 & 52.75 & 74.38 & 25.89 & 54.27 & 48.10 & 51.08 \\
0.00001 & 0.35 & 51.55 & 75.11 & 26.15 & 56.74 & 48.42 & 51.59 \\
0.00001 & 0.25 & 50.93 & 76.58 & 25.99 & 56.13 & 50.26 & 51.98 \\
0.00001 & 0.15 & 50.17 & 75.45 & 26.19 & 54.37 & 50.67 & 51.37 \\
\midrule
0.0001 & 0.45 & 50.90 & 77.56 & 25.83 & 54.33 & 48.90 & 51.50 \\
0.0001 & 0.35 & 51.20 & 75.67 & 26.65 & 57.21 & 48.78 & 51.90 \\
0.0001 & 0.25 & 50.32 & 76.64 & 27.09 & 56.36 & 50.47 & \textbf{52.18} \\
0.0001 & 0.15 & 50.09 & 74.79 & 25.27 & 55.21 & 51.82 & 51.44 \\
\midrule
0.001 & 0.15 & 49.05 & 76.03 & 26.36 & 54.85 & 51.49 & 51.56 \\
0.001 & 0.25 & 48.96 & 76.50 & 28.68 & 56.35 & 49.66 & 52.03 \\
0.001 & 0.35 & 51.25 & 74.41 & 26.51 & 56.58 & 50.05 & 51.76 \\
0.001 & 0.45 & 50.69 & 74.50 & 25.98 & 56.97 & 48.46 & 51.32 \\
\midrule
0.01 & 0.15 & 50.46 & 75.24 & 26.12 & 54.17 & 50.95 & 51.39 \\
0.01 & 0.25 & 51.02 & 76.28 & 27.75 & 55.48 & 49.93 & 52.09 \\
0.01 & 0.35 & 52.78 & 74.44 & 25.58 & 55.92 & 48.30 & 51.40 \\
0.01 & 0.45 & 50.09 & 74.87 & 27.60 & 54.69 & 48.68 & 51.19 \\
\bottomrule

\end{tabular}
\end{table}

\subsection{$\lambda(step)$ vs Constant $\lambda$} \label{app:lambda}
In this section, we compare our adaptive function $\lambda(step)$ against using a constant value for $\lambda$. To ensure a fair comparison, we conducted extensive experiments on LLaMa-3.2-3B, evaluating a wide range of constant values. Table~\ref{tab:lambda_selection} presents the results across different constant settings, demonstrating that even the best-performing constant value is outperformed by our adaptive $\lambda(step)$ approach.

\begin{table}[ht]
\caption{$\lambda(step)$ selection experiments on LLaMA-3.2-3B with fixed $\rho=0.7$.}
\label{tab:lambda_selection}
\centering
\small
\renewcommand{\arraystretch}{1.1}
\begin{tabular}{ccccccc}
\toprule
$\lambda$ & TruthfulQA & BoolQ & LogiQA & TydiQA & ASDiV & AVG \\
\midrule
constant value & 47.85 & 76.53 & 25.19 & 50.41 & 49.27 & 49.85 \\
$(t_{\max} - t_{\min}) \cdot \frac{\text{step}}{\text{total\_steps}}$ (linear) & \textbf{50.32} & \textbf{76.64} & \textbf{27.09} & \textbf{56.36} & \textbf{50.47} & \textbf{52.18} \\
\bottomrule
\end{tabular}
\end{table}

\subsection{Computational Cost}
\label{app:time_complexity}

As shown in the Table~\ref{tab:training_time}, we conducted a detailed analysis of computational cost. Given the consistent and significant performance improvements across different scales, this computational overhead is reasonable and justified. The preprocessing phase (warmup + influence computation) is a one-time cost that can be ignored if multiple training runs are conducted.

\begin{table}[ht]
	\caption{Training time comparison of different strategies for 3B and 8B models.}
	\label{tab:training_time}
	\centering
	\renewcommand{\arraystretch}{1.3}
	\begin{tabular}{lcc}
		\toprule[1.5pt]
		Component & 3B Model & 8B Model \\
		\midrule[1.5pt]
		
		\multicolumn{3}{c}{Preprocessing Phase} \\
		\midrule
		Warmup Training & 36 min & 52 min \\
		Influence Score Computation & 47 min & 73 min \\
		
		\midrule[1.5pt]
		\multicolumn{3}{c}{Training Phase} \\
		\midrule
		Full Token & 205 min & 347 min \\
		Ignoring & 166 min & 282 min \\
		Forgetting & 187 min & 311 min \\
		
		\midrule[1.5pt]
		\multicolumn{3}{c}{Total Time} \\
		\midrule
		Full Token (training only) & 205 min & 347 min \\
		Ignoring (warmup + score + training) & 249 min & 407 min \\
		Forgetting (warmup + score + training) & 270 min & 436 min \\
		
		\midrule[1.5pt]
		\multicolumn{3}{c}{Overhead} \\
		\midrule
		Ignoring vs Full Token & +21.46\% & +17.29\% \\
		Forgetting vs Full Token & +31.71\% & +25.65\% \\
		Forgetting vs Ignoring & +8.43\% & +7.12\% \\
		
		\bottomrule[1.5pt]
	\end{tabular}
\end{table}

\end{document}